\pdfoutput=1

\documentclass[11pt]{article}

\usepackage[]{EACL2023}

\usepackage{threeparttable}
\usepackage{times}
\usepackage{latexsym}
\definecolor{darkblue}{rgb}{0.0, 0.0, 0.55}
\hypersetup{colorlinks=true, citecolor=darkblue, linkcolor=darkblue, urlcolor=darkblue}
\definecolor{OliveGreen}{RGB}{0,153,0}
\definecolor{Mahogany}{rgb}{0.75, 0.25, 0.0}
\definecolor{RoyalPurple}{rgb}{0.47, 0.32, 0.66}
\definecolor{Purple}{rgb}{0.47, 0.32, 0.66}
\definecolor{ForestGreen}{rgb}{0.13, 0.55, 0.13}
\definecolor{Red}{rgb}{1.0, 0, 0}
\usepackage{url}
\usepackage{soul}
\usepackage{booktabs}
\usepackage{fontawesome}
\usepackage{graphicx}
\usepackage{tabularx} 
\usepackage{mathtools}
\usepackage{longtable}
\usepackage{adjustbox}
\usepackage[svgnames]{xcolor}
\usepackage{makecell}
\usepackage{booktabs}
\usepackage{bbm}
\usepackage{lineno}
\usepackage{tablefootnote}
\usepackage{color} 
\usepackage{soul} 
\usepackage{amsmath}
\usepackage{booktabs}
\usepackage{multirow}
\usepackage{tablefootnote}
\usepackage{wrapfig}
\usepackage{comment}
\newcommand{\tsbar}[1]{\textcolor{black}{\rule{#1}{1.6ex}}}        
\newcommand{\tfbar}[1]{\textcolor{black!50}{\rule{#1}{1.6ex}}}     
\usepackage{tikz}
\usetikzlibrary{patterns}

\newcommand{\otherbar}[1]{%
  \begin{tikzpicture}[baseline]
    \fill[pattern=north east lines, pattern color=black!40] (0,0) rectangle (#1,1.6ex);
    \draw[black!40] (0,0) rectangle (#1,1.6ex); 
  \end{tikzpicture}%
}
\usepackage[colorinlistoftodos,prependcaption,textsize=tiny]{todonotes}

\newcommand{\DALI}{\texttt{DALI}}
\newcommand{\DALIStrong}{\texttt{DALI}_{st}}
\newcommand{\MEXA}{\texttt{MEXA}_{F}}
\newcommand{\MEXAtask}{\texttt{MEXA}_{T}}

\usepackage{subcaption}
\usepackage{float}
\usepackage{placeins}

\usepackage[T1]{fontenc}

\usepackage[utf8]{inputenc}

\usepackage{microtype}

\usepackage{inconsolata}

\title{Can you map it to English? The Role of Cross-Lingual Alignment in the Multilingual Performance of LLMs}

\author{
\textbf{Kartik Ravisankar} \quad
\textbf{HyoJung Han} \quad
\textbf{Sarah Wiegreffe}\quad
\textbf{Marine Carpuat}\\[0.3em]
University of Maryland, College Park, MD, USA \\[0.3em]
\texttt{\{kravisan, hjhan, sarahwie, marine\}@umd.edu}
}

\begin{document}
\maketitle
\begin{abstract}
Large language models (LLMs) can answer prompts in many languages, despite being trained predominantly on English; yet, the mechanisms driving this generalization remain poorly understood. This work asks: How does an LLM's ability to align representations of non-English inputs to English impact its performance on natural language understanding (NLU) tasks? We study the role of representation alignment in instance-level task decisions, complementing prior analyses conducted both at the language level and task-independently. We introduce the Discriminative Alignment Index ($\DALI$) to quantify instance-level alignment across 24 languages other than English and three distinct NLU tasks. Results show that incorrect NLU predictions are strongly associated with lower representation alignment with English in the model's middle layers. Through activation patching, we show that incorrect predictions in languages other than English can be fixed by patching their parallel English activations in the middle layers, thereby demonstrating the causal role of representation (mis)alignment in cross-lingual correctness.\footnote{{\faGithub\textbf{ Code: }\href{https://github.com/Kartik21/XLingAlignment}{github.com/Kartik21/XLingAlignment}}}
\end{abstract}

\section{Introduction}
\label{sec: Introduction}
Large language models (LLMs) exhibit impressive multilingual capabilities, successfully performing natural language understanding (NLU) tasks such as reading comprehension and common-sense reasoning in languages other than English despite being pre-trained mostly on English text \cite{touvron2023llama2openfoundation,muennighoff-etal-2023-crosslingual}.
This ability to transfer NLU capabilities from high-resource languages (e.g., English) to lower-resource ones has been well-documented in encoder-only architectures \citep{conneau-etal-2018-xnli,conneau-etal-2020-unsupervised,yang-etal-2019-paws,devlin-etal-2019-bert}. However, the capacity of decoder-only LLMs to internalize and transfer knowledge across languages remains relatively underexplored \citep{hammerl-etal-2024-understanding}.

Recent work has focused on understanding how LLMs predominantly trained in English process multilingual text. \citet{wendler-etal-2024-llamas} analyze intermediate representations in Llama-2~\citep{touvron2023llama2openfoundation} through early exit strategies \citep{nostalgebraist}, demonstrating an implicit pivot through English in the middle layers while processing non-English text. This raises the question of whether an LLM's ability to align representations of non-English text to its corresponding parallel English text representations is predictive of its capabilities in languages other than English. 

Prior work shows that representational alignment measured using independent parallel corpora correlates with multilingual task accuracy: \citet{kargaran-etal-2025-mexa} introduced \texttt{MEXA}, a cross-lingual retrieval-based score for languages other than English, computed by comparing non-English representations to English representations from the same model across a fixed set of parallel texts. The authors demonstrate that \texttt{MEXA} scores are strongly correlated with multilingual task accuracy (i.e., languages with better \texttt{MEXA} scores exhibit better task accuracy), suggesting that cross-lingual alignment is a good indicator of an LLM's multilingual capability. However, language-level correlational studies analyze alignment at the aggregate level, using a corpus independent of the NLU task, leaving open the question of how representation alignment affects model behavior on individual NLU instances.

\begin{figure*}[ht]
    \centering
    \includegraphics[width=0.73\linewidth]{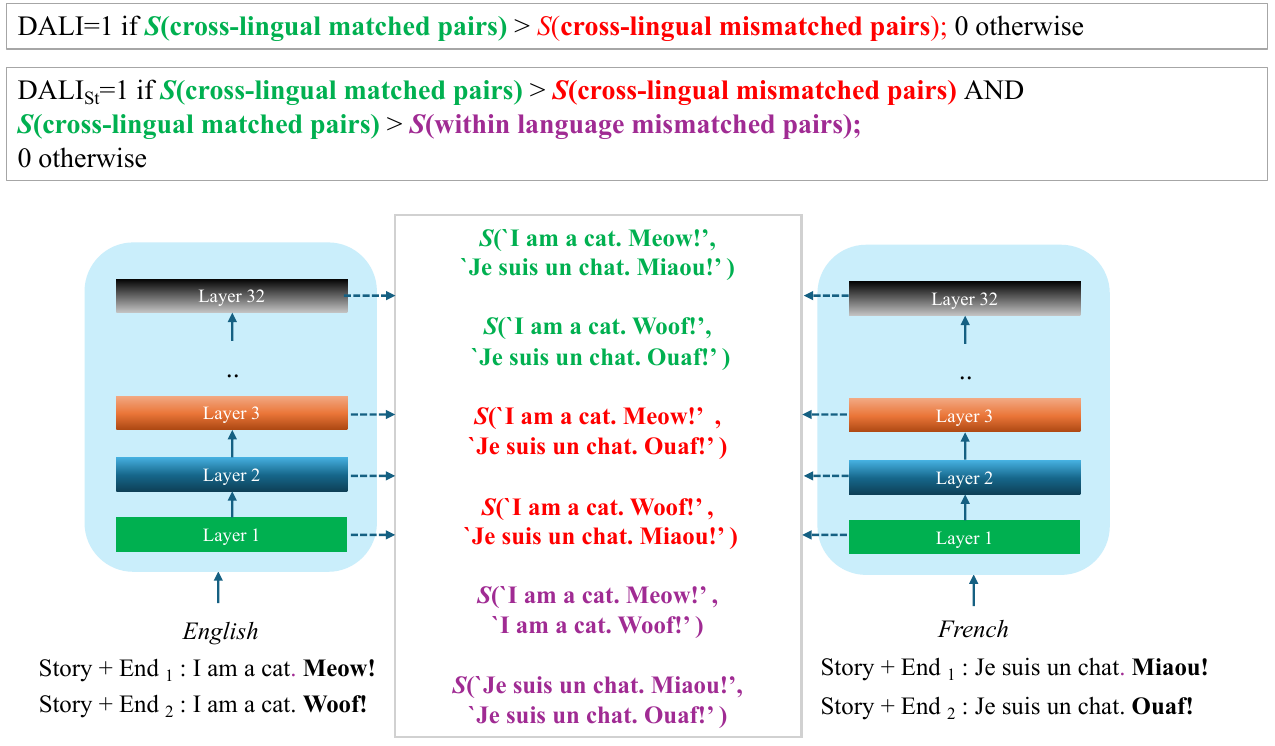}
        \caption{\texttt{DALI} is calculated at the instance-level across transformer layers using its representations. The model is tasked with picking the right ending (\textit{`Meow/Woof'} in English; \textit{`Miaou/Ouaf'} in French) given a premise (\textit{`I am a cat/Je suis un chat.'} in English and French, respectively). \texttt{DALI}=1 if the cosine similarity $S$ of both the \textcolor{OliveGreen}{cross-lingual matched pairs} exceeds both the \textcolor{Red}{mismatched pairs}, indicating the ability of the model to distinguish semantically equivalent English and French text from non-equivalent representation pairs. A stricter variant, $\DALIStrong$ adds an additional condition that $S$ of \textcolor{OliveGreen}{cross-lingual matched pairs} must exceed the \textcolor{Purple}{intra-lingual mismatched pairs}.}
        \label{fig:DALI_illustration}
\end{figure*}

\begin{figure}[ht]
    \centering
 \includegraphics[width=\linewidth]{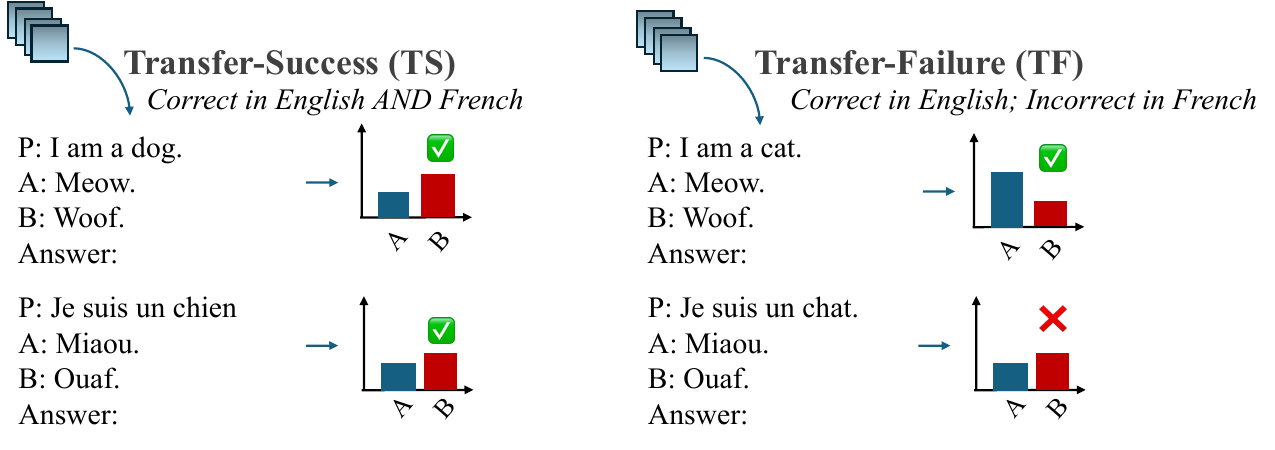}
        \caption{French TS and TF samples in MCQA format. }
        \label{fig:TS_TF}
\end{figure}

Our work seeks to address this gap and determine whether LLMs make better multilingual predictions when given inputs well-aligned with English. We ask two research questions (RQ):
\begin{enumerate}
\item \textbf{RQ1: }Is cross-lingual alignment associated with task accuracy at the instance-level of a specific NLU task within a language?
\item \textbf{RQ2: }Does misalignment with English representations causally explain failures on inputs in other languages?

\end{enumerate}
Following \citet{kargaran-etal-2025-mexa}, we focus on multilingual discriminative NLU tasks. To answer RQ1, we introduce the Discriminative Alignment Index score ($\DALI$) and its variant ($\DALIStrong$), which quantify representation alignment at the instance level (\autoref{fig:DALI_illustration}). 
By establishing instance-level metrics to quantify alignment, we compare two groups in which cross-lingual transfer from English yields distinct outcomes: \textbf{Transfer Success (TS)} and \textbf{Transfer Failure (TF)} (\autoref{fig:TS_TF}). We find that TS instances are consistently more aligned with their English counterparts than TF instances across benchmarks, languages, and models.

We answer RQ2 through controlled activation patching (\autoref{fig:Patching_illustration}): we patch semantically equivalent English representations at layer $\lambda$ onto the corresponding non-English forward pass of TF instances. We then observe whether the patch is more successful at flipping the model's prediction to the correct answer than a patch from a control (i.e., semantically non-equivalent) instance. Our experiments reveal that semantically equivalent English patches are consistently more effective than control patches at correcting predictions, with effects concentrated in specific intermediate layers, providing causal evidence that alignment with English representations drives successful cross-lingual transfer. 


\section{Methods}

We first introduce metrics to quantify representation alignment with English at an instance level, and compare alignment between TS and TF instances (\S\ref{sec:xling_alignment}). Then, we formalize our activation patching setup, which establishes a causal link between alignment and cross-lingual transfer (\S\ref{sec:activation_patching}).

\subsection{Preliminaries}
\texttt{MEXA} \citep{kargaran-etal-2025-mexa}, henceforth denoted as $\MEXA$, measures a model's general cross-lingual alignment ability with English using a fixed set of sentences from parallel datasets such as FLORES-200 \citep{nllbteam2022languageleftbehindscaling}. $\MEXA$ follows the concept of weak alignment \citep{hammerl-etal-2024-understanding} defined as the proportion of samples that are \textit{`aligned'}. For languages $\text{L}_1,\text{L}_2$ drawn from the set $\mathcal{L}$, let $(u_i,v_i)$ be pairs of sentence representations derived from the layer $\lambda$ of a transformer where $i=1,\ldots,N; u \in \text{L}_1, v\in \text{L}_2$. We say a sample is \textit{`aligned'} if it has a higher cosine-similarity with its parallel instance than with other non-parallel instances. Formally, we define $\MEXA$ as follows:

{\small
\begin{equation}
\label{eq:MEXA}
\begin{aligned}
{\MEXA}_{(\text{L}_1,\text{L}_2, \lambda)} &= \frac{1}{N}\sum_{i=1}^{N}{1}\bigl(S(u_i,v_i) > \\
&\quad \max_{j\in 1,\ldots N; j\neq i}\bigl(\{S(u_i,v_j)\}\cup\{S(u_j, v_i)\}\bigr)\bigr)
\end{aligned}
\end{equation}
}

Since transformer representations exhibit anisotropy \citep{ethayarajh-2019-contextual}, i.e., they occupy a narrow, directional cone in the latent space rather than being uniformly distributed, cosine similarity scores are high even for semantically unrelated text, making it challenging to distinguish genuine alignment from spurious directional clustering. By assigning a binary score instead of using raw cosine similarities, $\MEXA$ mitigates this issue. Any parallel dataset can be used to compute $\MEXA$, as evidenced by the original study, which also used the Bible \citep{mayer-cysouw-2014-creating} corpus in addition to FLORES-200. Empirically, $\text{max}_{\lambda} \bigl\{{\MEXA}_{(\text{L}_1=\text{English},\text{L}_2)}\bigr\}$ is strongly correlated with NLU model performance in $\text{L}_2$, such that languages with higher $\MEXA$ scores tend to achieve higher task accuracy and vice versa.

\subsection{Instance-level metrics for Cross-lingual Alignment}
\label{sec:xling_alignment}
While $\MEXA$ provides a general magnitude of cross-lingual alignment at the language-level, our objective is to quantify cross-lingual alignment at an instance-level on a discriminative task. To this end, we propose three metrics:  1) $\DALI$, 2) $\DALIStrong$ - a stricter variant of $\DALI$, and 3) $\MEXAtask$ - a task-specific variant of $\MEXA$. Although these metrics can be applied to any two languages, we fix $\text{L}_1 = \text{English}$ for all our analyses, as we are interested in examining the LLM's ability to align non-English representations with their corresponding English counterparts. 
\paragraph{$\DALI$}
Consider a discriminative task $\mathcal{D}$ across multiple languages, where each instance has a premise $P$ and string answer options $o_1,o_2,\ldots,o_{n_{\text{opt}}}$. The model is tasked with picking the correct option. \autoref{fig:DALI_illustration} presents one example where the model is given a premise in both English ($P_{\textit{Eng}}$ = \textit{`I am a cat.'}) and French ($P_{\textit{Fr}}$ = \textit{`Je suis un chat.'}), and $n_{\text{opt}}=2$. The model is tasked with picking the right ending among the options for the given premise, conditioned on the language of the premise: in English, $o_{1,\text{Eng}}$ = \textit{Meow!} and $o_{2,{\text{Eng}}}$ = \textit{Woof!}; in French, $o_{1,{\text{Fr}}}$ = \textit{Miaou!} and $o_{2,{\text{Fr}}}$ = \textit{Ouaf!}.

We extract the representations of the premise-ending combinations ($P+o_1, P+o_2$) in both languages from various layers $\lambda$ of an LM. We set $\DALI$ = 1 if the similarity ($S$) of \textcolor{OliveGreen}{cross-lingual matched pairs} exceeds \textcolor{Red}{mismatched pairs}. In principle, we can use any vector similarity metric, but we use cosine similarity as a representative option. $\DALI$  thus captures the model's ability to align parallel premise and ending representations of English and non-English samples. We define $\DALI$ for a given sample across languages $\text{L}_1,\text{L}_2$ based on the representations in layer $\lambda$ of a transformer as follows:

{\small
\begin{equation} \label{eq:DALI} 
\begin{split} \DALI_{\text{L}_{1}, \text{L}_{2},\lambda} = \begin{cases} 1,\text{if }S \bigl((P+\text{o}_{i})_{\text{L}_{1}}, (P+\text{o}_{i})_{\text{L}_{2}}\bigr) \\
\quad > S \bigl((P+\text{o}_{i})_{\text{L}_{1}}, (P+\text{o}_{j})_{\text{L}_{2}}\bigr) \\  \quad \forall i,j= 1,\ldots,n_{\text{opt}}; \, i \neq j \\ 0, \text{otherwise} \end{cases} \end{split} 
\end{equation}
}

We follow the same approach as \texttt{MEXA} \citep{kargaran-etal-2025-mexa} by assigning a binary $\DALI$ score per sample instead of cosine similarities. However, the small pool of mismatched pairs reduces $\DALI$'s discriminative power: for instance, a 2-option task involves only two cross-lingual mismatches, increasing the likelihood of false positives.

\paragraph{$\DALIStrong$}
To address this issue, we introduce a stricter metric, $\DALIStrong$. We enforce an additional criterion that the cosine similarity of cross-lingual matched pairs must surpass all \textcolor{Purple}{within-language mismatched pairs} (see \autoref{fig:DALI_illustration}). This imposes a stricter threshold by capturing whether the similarity between cross-lingual pairs is higher than that of non-equivalent sentences within the same language. We formally define $\DALIStrong$ as follows:

{\small 
\begin{equation} \label{eq:DALIstrong} 
\begin{split} {\DALIStrong}_{,\text{L}_{1}, \text{L}_{2},\lambda} = \begin{cases} 1,\text{if $\DALI=1$ and }\\
S \bigl((P+\text{o}_{i})_{\text{L}_{1}}, (P+\text{o}_{i})_{\text{L}_{2}}\bigr) \\
\quad > S \bigl((P+\text{o}_{i})_{\text{L}_{k}}, (P+\text{o}_{j})_{\text{L}_{k}}\bigr) \\  \quad \forall i,j= 1,\ldots,n_{\text{opt}};i \neq j, \forall k=1,2 \\ 0, \text{otherwise} \end{cases} \end{split} 
\end{equation}
}

\paragraph{$\MEXAtask$}
\label{ref:sec:MEXAtask}
We also compute a task-specific version of $\MEXA$ (\autoref{eq:MEXA}) that allows for instance-level alignment measurements for each NLU sample. The only difference in calculating $\MEXAtask$ is that $u \in P_{\text{Eng}}, v\in P_{\text{L}_2}$ from task $\mathcal{D}$ as opposed to being generic sentences from the FLORES dataset. 

\paragraph{Differences between the metrics}
\label{sec:differencesinalignment}
While the three proposed metrics all quantify cross-lingual alignment at the instance level, they differ in their definitions. $\DALI$ and $\DALIStrong$ ensure that premise+option pairs in a given NLU instance are aligned, meaning the contrastive examples are within a given test instance. $\MEXAtask$, on the other hand, focuses only on whether the representations of parallel premises across languages are more aligned than non-parallel premises. Based on \autoref{fig:DALI_illustration}'s example, 
\begin{equation}
\small
\MEXAtask = 
\begin{cases}
1 & \text{if } S(\text{`Je suis un chat'}, \text{`I am a cat'}) \\
  & \quad > \max \big\{ S(\text{`I am a cat'}, \text{{other  ${P}_{\text{Fr}}$}}), \\
  & \qquad S(\text{`Je suis un chat'}, \text{other ${P}_{\text{Eng}}$}) \big\}
\end{cases}
\end{equation}
Due to the number of contrastive parallel samples ($2N-2$ if there are $N$ total samples), the probability of $\MEXAtask=1$ occurring by chance is low. On the other hand, $\DALI$ relies on within-sample mismatched pairs, which are inherently limited by task design: a 2-option task involves only two mismatched cross-lingual pairs. While $\DALIStrong$ mitigates this by enforcing a stricter criterion, tasks with few options remain vulnerable to false positives due to anisotropy ($\S$\ref{sec:Qualitativeanalysis}).
\begin{table}[t!]
\small
\centering
\begin{tabular}{p{2.7cm}p{1.7cm}m{0.3cm}<{\centering}m{0.3cm}<{\centering}m{0.3cm}<{\centering}}
\hline
\textbf{Benchmark} & \textbf{Description} & $|\mathcal{L}|$ & $n_\textbf{opt}$ & $N$ \\
\hline
\makecell[l]{Belebele \\ \footnotesize{\citep{bandarkar-etal-2024-belebele}}} 
  & \makecell[l]{Reading\\comprehension} 
  & 122 & 4 & 900\\
\makecell[l]{Xstorycloze \\ \footnotesize{\citep{lin-etal-2022-shot}}} 
  & \makecell[l]{Narrative\\understanding } 
  & 11 & 2 & 1511\\
\makecell[l]{Xcopa \\ \footnotesize{\citep{ponti-etal-2020-xcopa}}} 
  & \makecell[l]{Commonsense\\ reasoning } 
  & 11 & 2 & 500\\
\hline
\end{tabular}
\caption{NLU discriminative benchmarks-- Number of languages ($|\mathcal{L}|$), options per sample ($n_{\text{opt}}$), and number of samples per language $(N)$.}
\label{tab:disc_benchmarks}
\end{table}
\paragraph{Associative analysis of cross-lingual alignment}
\label{sec:xlingalignment_analysis}
We analyze the role of alignment between \textbf{TS} and \textbf{TF} instances for each language.  We hypothesize that TS samples exhibit \emph{higher} cross-lingual alignment than TF samples, and we test this hypothesis statistically. It is worth noting that $\DALI$, $\DALIStrong$, and $\MEXAtask$ are derived across all transformer layers for each sample, producing a binary vector (of the same length as the number of layers) per instance. We compute the \% of samples with alignment=1 at each layer $\lambda$ and identify the layer with the largest alignment overall (denoted as $\lambda_{max}$). Although statistical tests between alignment measures can be conducted at any layer, we localize our tests to the layer with the highest alignment for each metric, as this is where the model's cross-lingual alignment mechanism is most actively and successfully engaged. We compare alignment (\% alignment=1 at $\lambda_{max}$) across the two groups using a z-test for proportions against a one-sided alternative that \% aligned in TS samples exceeds \% aligned in TF samples at a level $(\alpha)$ of 0.05. A positive $\Delta_{\text{TS-TF}}$(alignment) means that samples that generalize well have a higher degree of alignment than samples that do not.
\begin{figure}[htbp]
    \centering
    \includegraphics[width=0.8\linewidth]{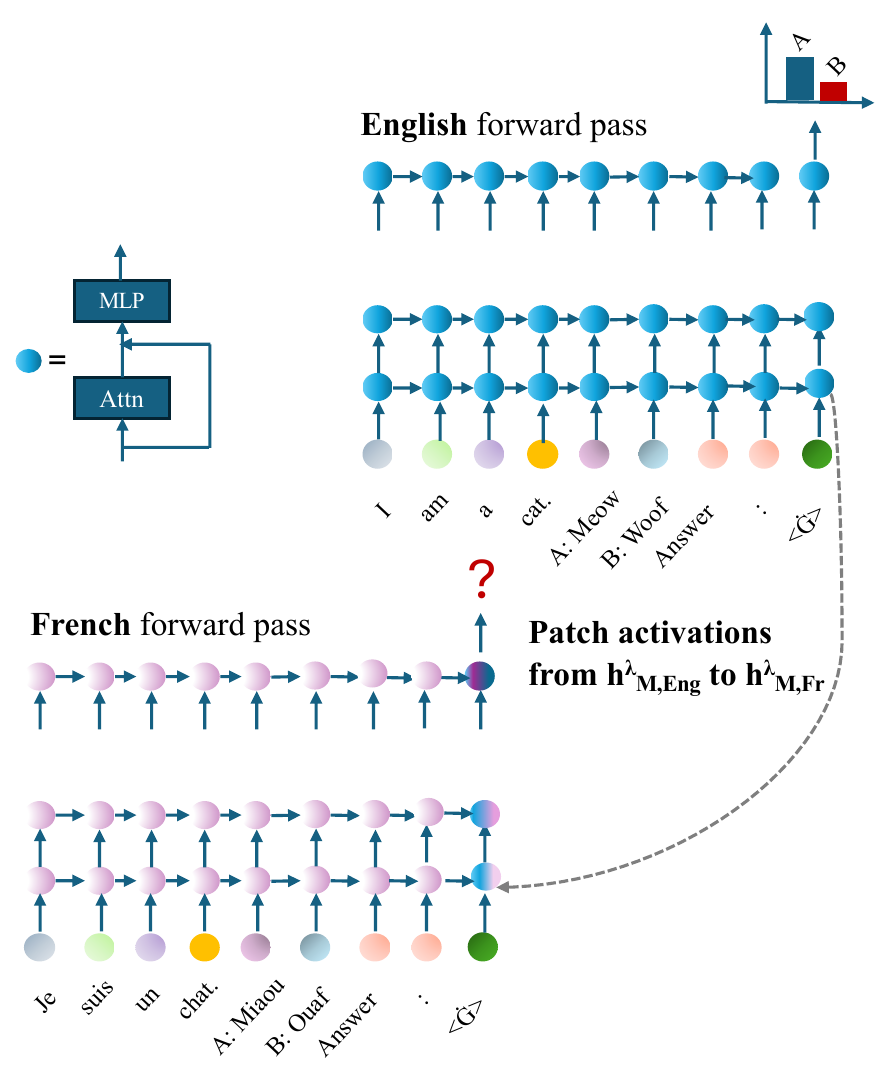}
        \caption{We analyze the causal effect of alignment on task accuracy by patching a successful English forward pass $h_{M,Eng}^{\lambda}$ to the unsuccessful French forward pass at layer $\lambda$ ($h_{M,Fr}^{\lambda}$), thereby treating $h_{Eng}^{\lambda}$ as a causal mediator for the French sample's success. We measure the effectiveness of the patch by tracking the \% of samples that flip to the correct prediction upon patching. }
\label{fig:Patching_illustration}
\end{figure}
\subsection{Activation Patching}
\label{sec:activation_patching}
To determine whether non-English failures stem from the model's inability to construct representations equivalent to those produced by the English forward pass, we perform activation patching. Activation patching is an interpretability technique \citep{vig2020investigating, meng2022locating} that allows one to test whether an LM's representation at layer $\lambda$ ($h^{\lambda}$) is a causal mediator \citep{DBLP:journals/corr/abs-2408-01416} for the model's behavior. 
Transformer architectures process input tokens\footnote{Without loss of generality, we assume the input tokens are specific to language $L$. } $x_1, x_2,\ldots,x_m,\ldots, x_M \in \mathcal{V}$ and transform each token $x_m$ into the $ d$-dimensional embedding space at layer $0$ ($h_m^{0}$). For each token and layers $\lambda \in \{1,\ldots, \Lambda\}$, the transformer model updates the residual stream in the following way, where $f_\lambda(.)$ refers to the self-attention block followed by the multi-layer perceptron (MLP) block at layer $\lambda$:
\begin{equation}
\label{eq:residstream}
h_{m,L}^{\lambda} = h_{m,L}^{(\lambda-1)}+f_{\lambda}(h_{1,L}^{(\lambda-1)},\ldots,h_{m,L}^{(\lambda-1)})
\end{equation}
The next-token logits are derived by unembedding the vector output by the final layer, $h_{m,L}^{\Lambda}$, back to the vocabulary space $\mathcal{V}$. For simplicity, we drop the token index subscript $m$ in subsequent sections, noting that patching is performed only at specific positions such as the last or penultimate token.

We run two forward passes for each TF sample: one in English, which the model answers correctly, and one in another language, which the model answers incorrectly. If the non-English forward pass had constructed a representation at layer $\lambda$ equivalent to $h_{\text{Eng}}^{\lambda}$, would the non-English prediction have instead been correct? We test this by copying the activations at specific token positions from the English forward pass ($h_{\text{Eng}}^{\lambda}$), and patching them onto the corresponding non-English forward pass at the same layer ($h_{\text{X}}^{\lambda}$) (see \autoref{fig:Patching_illustration}). We iterate this process across all layers of the transformer, enabling us to localize the layers where aligned representations are sufficient to correct failures in other languages.

Unlike in \S\ref{sec:xling_alignment}'s experiments, which computed alignment by concatenating the premise and options, we frame the discriminative NLU tasks as Multiple-Choice Question Answering (MCQA) to ensure that key token positions align across the two instances (a prerequisite for patching). Formally, we have a set of target tokens $\mathcal{Y}=\{y_{1}, y_{2}, \dots, y_{n_{\text{opt}}}\} \text{ (e.g., } \{\mathtt{A}, \mathtt{B}\} $ in the 2-choice MCQA task) that represent string answer choices $o_1,o_2,\ldots,o_{n_{\text{opt}}}$ for each instance. We use the same Latin-script target tokens across languages. The model's prediction is considered correct if the logit for the target token representing the correct answer ($y_c$) is higher than those for the other target tokens (Refer \autoref{fig:TS_TF} for an example).

\begin{figure}[h]
    \centering
\includegraphics[width=\linewidth]{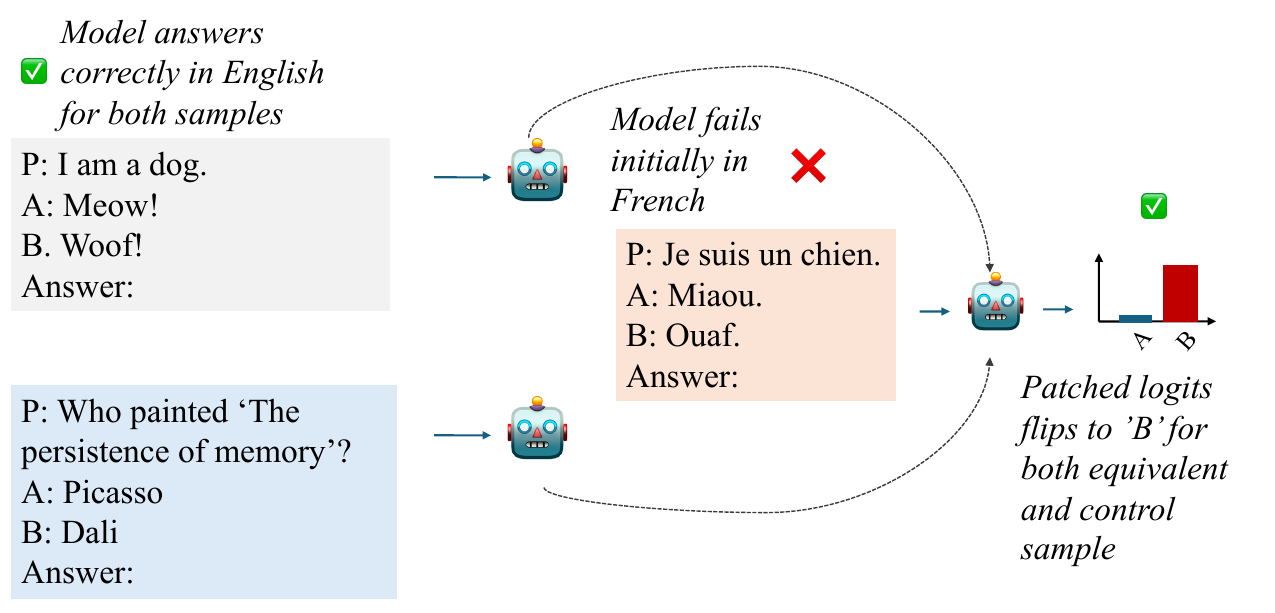}
\caption{Control experiment to check if $h_{\text{Eng}}^{\lambda}$ encodes semantic concepts or just the target token.}
\label{fig:control_patching}
\end{figure}

\paragraph{Control}
In an MCQA setting, models may encode either the predicted answer choice string or its corresponding answer symbol token (or both) in their hidden states, depending on the layer \citep{wiegreffe2025answerassembleaceunderstanding}. If $h^{\lambda}_{\text{Eng}}$ encoded the target token prediction rather than \emph{only} information specific to the instance, an unrelated English sample with the same $y_c$ (e.g., `\texttt{A}') would be equally effective at correcting the non-English prediction. To isolate the layers and token positions where $h_{\text{Eng}}^{\lambda}$  encodes concepts or semantics and not simply the target token or pointers to it, we also perform control patching where we copy activations associated with unrelated English samples, but with the same $y_c$ as the current instance pair. In \autoref{fig:control_patching}, we demonstrate patching $h_{\text{Eng}}^{\lambda}$ from an unrelated sample with $y_{c}=$ `\texttt{B}' to repair the French sample. The successful control patch demonstrates that $h_{\text{Eng}}^{\lambda}$ cannot be interpreted as encoding \emph{only} semantic information and is thus inconclusive for testing cross-lingual alignment for this particular instance. The control experiment exemplifies why we frame the tasks as MCQA: it provides the flexibility to evaluate and identify key, consistent token positions despite differences in tokenization across languages. 
\paragraph{Token position} 
A key facet of our experiment is the token position $m$ from which we must copy our English activations and patch them into the non-English forward pass. We consider two positions in our Multiple-Choice Question Answering (MCQA) prompts—at the \textbf{penultimate token} ($M-1$) and the \textbf{last token} ($M$), since both positions capture the task logic of the entire sequence across languages.

\paragraph{Metrics}
We evaluate the causal effectiveness of English representations on non-English instances in three dimensions: 
\begin{enumerate}
    \item \textbf{Mean logits} over the non-English TF samples, identifying the layer $\lambda$ where the logit associated with the gold token ($y_c$) exceeds the logits assigned to the distractor target tokens,
    \item \textbf{\% flips}, defined as the proportion of non-English TF samples where the $y_c$'s logit exceeds the distractor logits post-patching, including the logit of the incorrect token initially predicted pre-patching, and
    \item \textbf{$\Delta$ \% flips} between equivalent and control patching setups, since this isolates the role of conceptual alignment from simple target-token production. 
\end{enumerate}

\section{Experimental Setup}
\paragraph{Tasks}
\label{sec:task-acc}
We evaluate LLMs on three multilingual NLU benchmarks (\autoref{tab:disc_benchmarks}). Our study is made possible by their parallel nature, in which premise-option pairs are structurally identical and equivalent across languages (e.g., `I am a cat' in English and `Je suis un chat' in French).  

The specific prompts used are detailed in \autoref{fig:evaluation_prompts}. For each sample, we use the MCQA format and evaluate the logits for each target token $y \in \mathcal{Y}$. We consider the model's decision correct if the gold token ($y_{\text{c}}$) receives the highest logits. We label an instance as TS if the model is correct on both the English and the non-English NLU sample, and TF if it is correct only on the English version (\autoref{fig:TS_TF}).

\begin{figure*}[htbp]
\centering
\includegraphics[width=0.85\linewidth]{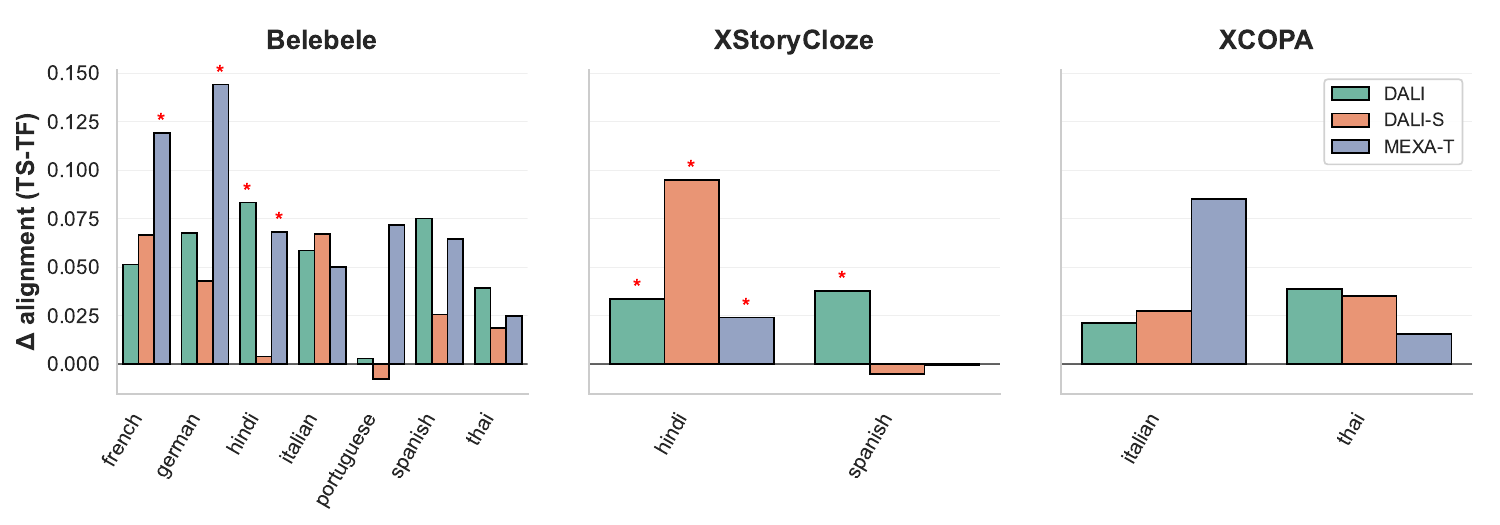}
\caption{$\Delta$ Alignment between TS and TF samples using $\DALI$, $\DALIStrong$, and $\MEXAtask$ across the three benchmarks in Llama 3.1 8B. Positive $\Delta$s indicate that alignment is higher in TS samples than
TF samples for a given language. The asterisk (*) indicates that this difference is statistically significant (p < 0.05).}
\label{fig:alignmentresults_Llama3.1_main}
\end{figure*}
\paragraph{Models and Languages}
We conduct our alignment association experiments across three diverse open-source models, namely Llama3.1 8B , Llama3.1 8B (it) \citep{grattafiori2024llama3herdmodels}, and Aya23 8B \citep{aryabumi2024aya23openweight}.  This selection encompasses diverse architectural approaches, supported languages, and training methodologies. Critically, these models report and document the range of languages they support, which enabled us to construct an intersection set with the languages available in individual benchmarks (\autoref{tab:models_lang}). 

Since the patching experiments are done layer-wise, this yields 2$\Lambda$ forward passes per TF sample in a given language. Due to computational constraints, we limit our patching experiments to the Llama3.1 8B model and the languages it natively supports.  We perform our patching experiments using \texttt{nnsight} \cite{fiottokaufman2025nnsightndifdemocratizingaccess}, a Python package used to access model internals.

\paragraph{Embeddings} Following prior work \citep{neelakantan2022textcodeembeddingscontrastive,wang-etal-2024-improving-text,kargaran-etal-2025-mexa,10.1609/aaai.v39i27.35038}, we extract the embeddings corresponding to the last token of the text ($h_M^\lambda$) across each layer of the transformer to calculate our instance-level alignment metrics. 
\section{Results}
We present the results of our instance-level alignment analysis ($\S$\ref{sec:xling_alignment_results}), the activation patching ($\S$\ref{sec:patching_results}), and the control patching experiments ($\S$\ref{sec:control_results}).

\begin{table}[htbp]
\centering
\small
\newlength{\barwidth}
\setlength{\barwidth}{2.0cm} 
\resizebox{\columnwidth}{!}{%
\begin{tabular}{llccl}
\toprule
Benchmark & Language & $n_{acc}$ & $\%_{acc}$ & TS/TF distribution \\
\midrule
\multirow{8}{*}{Belebele}
& English      & 723 & 80.3 &  \otherbar{0.803\barwidth} \\
& French       & 648 & 72.0 &  \tsbar{0.676\barwidth}\tfbar{0.128\barwidth}\\
& German       & 630 & 70.0 &  \tsbar{0.651\barwidth}\tfbar{0.152\barwidth} \\
& Hindi        & 499 & 55.4 &  \tsbar{0.518\barwidth}\tfbar{0.286\barwidth}\\
& Italian      & 624 & 69.3 &  \tsbar{0.639\barwidth}\tfbar{0.165\barwidth} \\
& Portuguese   & 655 & 72.8 &  \tsbar{0.682\barwidth}\tfbar{0.121\barwidth} \\
& Spanish      & 635 & 70.6 &  \tsbar{0.653\barwidth}\tfbar{0.15\barwidth} \\
& Thai         & 499 & 55.4 &  \tsbar{0.518\barwidth}\tfbar{0.286\barwidth} \\
\midrule
\multirow{3}{*}{Xstorycloze} 
& English      & 1451 & 96.0 & \otherbar{0.96\barwidth} \\
& Hindi        & 1263 & 83.6 & \tsbar{0.821\barwidth}\tfbar{0.139\barwidth} \\
& Spanish      & 1398 & 92.5 & \tsbar{0.903\barwidth}\tfbar{0.057\barwidth} \\
\midrule
\multirow{4}{*}{Xcopa} 
& English-Italian  & 429 & 85.8 & \otherbar{0.86\barwidth} \\
& Italian      & 408 & 81.6 & \tsbar{0.748\barwidth}\tfbar{0.11\barwidth} \\
& English-Thai & 356 & 71.2 & \otherbar{0.71\barwidth} \\
& Thai         & 288 & 57.6 & \tsbar{0.524\barwidth}\tfbar{0.188\barwidth} \\
\bottomrule
\end{tabular}%
}
\caption{Accuracy results for Llama3.1 8B. $n_{acc}$ and $\%_{acc}$ present the number of accurate instances in a given language. The distribution column presents the \% of accurate samples in English ( \protect\otherbar{0.6em} ) for each benchmark, as well as the breakdown of TS ( \protect\tsbar{0.6em} ) and TF ( \protect\tfbar{0.6em} ) instances for each language-benchmark combination. }
\label{tab:acc_stats}
\end{table}

\subsection{Is alignment stronger when transfer succeeds?}
\label{sec:xling_alignment_results}
\autoref{tab:acc_stats} presents the accuracies achieved by the Llama3.1 8B model for each language-benchmark combination, along with the \% of TS and TF samples used in our alignment experiments.

The $\Delta_{\text{TS-TF}}$(alignment) is consistently positive (\autoref{fig:alignmentresults_Llama3.1_main}) in 30 out of the 33 (language $\times$ benchmark $\times$ alignment metric) combinations, including 8 with statistically significant differences. This finding directly addresses RQ1 as we demonstrate an association between cross-lingual alignment with English and instance-level success. Considering the various aspects of cross-lingual alignment that these metrics measure ($\S$\ref{sec:differencesinalignment}), the consistently positive $\Delta$s point towards a strong association between superior representational alignment and successful transfer generalization. We observe similar positive results in other models (\autoref{fig:alignmentresults_appendix}).

\begin{figure*}[t]
    \centering
    \begin{subfigure}[b]{0.45\textwidth}
        \centering
        \includegraphics[width=0.9\textwidth]{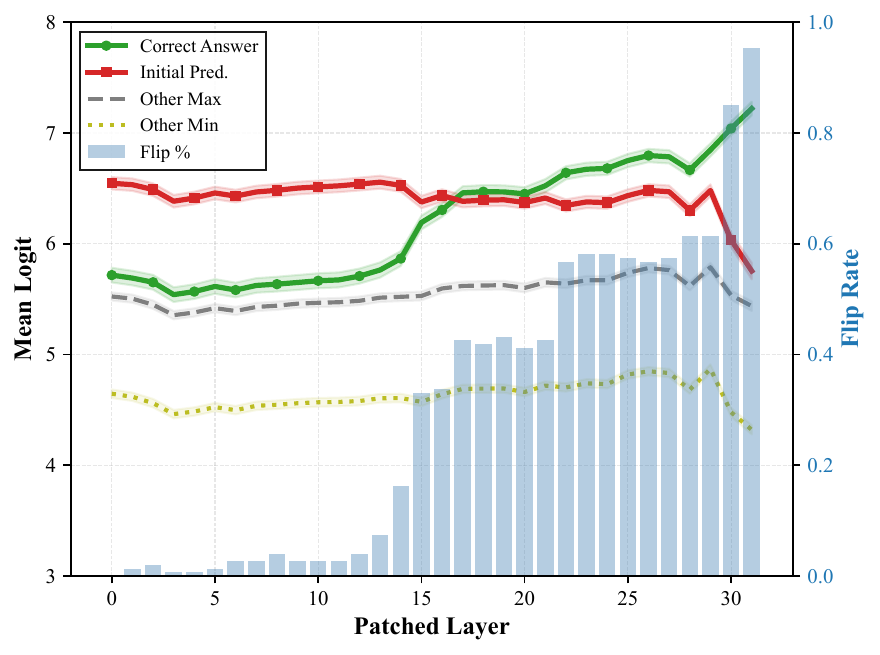}
        \caption{Belebele - Italian (last-token)}
        \label{fig:belebele_lasttokenpatch_main}
    \end{subfigure}
    \hfill
    \begin{subfigure}[b]{0.45\textwidth}
        \centering
        \includegraphics[width=0.9\textwidth]{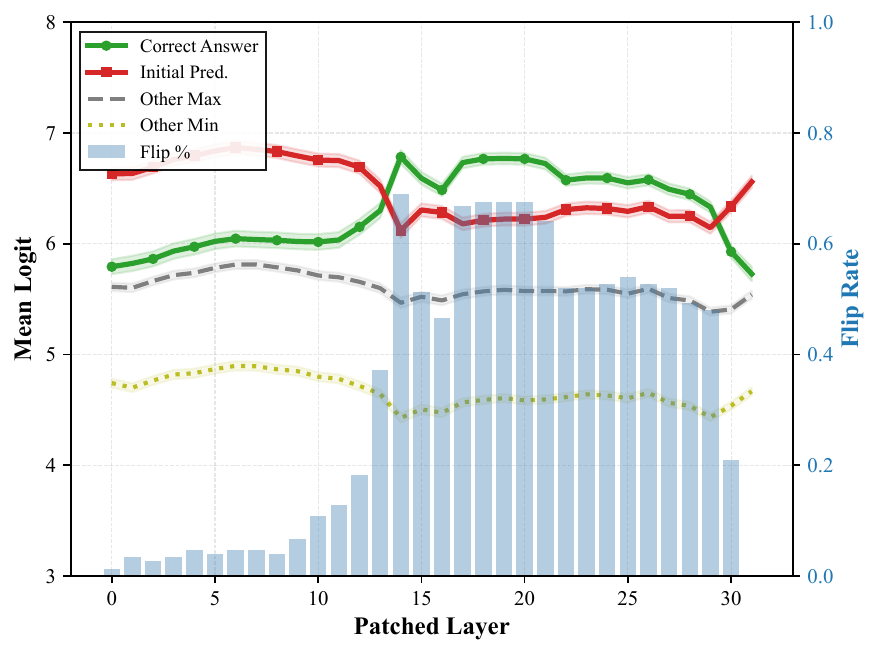}
        \caption{Belebele - Italian (penultimate token)}
        \label{fig:belebele_penultimatetokenpatch_main}
    \end{subfigure}
    \caption{Mean logit trajectory and \% flip for Italian TF samples (N=148) in the belebele benchmark: Blue bars represent the \% of TF samples that flip to the correct prediction; \textcolor{OliveGreen}{Green} line refers to the mean logits of the gold-token ($t_{c}$) and \textcolor{Red}{Red} line refers to the mean logits of the original incorrect prediction.}
    \label{fig:belebele_patching_main_results}
\end{figure*}
\subsection{Does patching $h_{\text{Eng}}^{\lambda}$ correct failures?}
\label{sec:patching_results}

We present the results of the Llama3.1 8B semantically equivalent patching experiment on the Belebele benchmark for Italian (148 TF samples) in the last token (\autoref{fig:belebele_lasttokenpatch_main}) and the penultimate token (\autoref{fig:belebele_penultimatetokenpatch_main}) settings. Results for other language-benchmark combinations are presented in Figures \ref{fig:patching_results_belebele_appendix}, \ref{fig:patching_results_xstorycloze_appendix}, and 
\ref{fig:patching_results_xcopa_appendix}. We divide the plots into three regions of the model's decision process. 
\paragraph{Last Token -}\textbf{Early layers (0-15): }\% flip remains low, and the mean logits of the \textcolor{Red}{initial incorrect answer} exceed the \textcolor{OliveGreen}{gold answer token}, indicating that $h_{\text{Eng}}^{\lambda}$ has not yet aggregated the right features necessary to overturn the wrong prediction; \textbf{Middle layers (16-25): }\% flip rises sharply in layer 16 and continues to rise until layer 25. This suggests that $h_{\text{Eng}}$ corresponding to the last token in the middle layers (as early as layer 17) encodes a language-agnostic representation that is sufficient to steer the non-English samples towards the gold token; \textbf{Later layers (25-32): }\% flip steadily increases followed by maximal correction, trivially reaching $\approx$ 100\% in the last layer (i.e., if $h_{\text{Eng}}^{\Lambda}$ is patched to $h_{\text{Ita}}^{\Lambda}$ at the last token, the model is essentially treating it as an English sample).

\paragraph{Penultimate Token - }\textbf{Early layers (0-13): }\% flip remains low, indicating that $h_{\text{Eng}}^{\lambda}$ has not yet aggregated the right features necessary to overcome the Italian failures; \textbf{Middle layers (14-25): }This is followed by a spike \% flip in layer 14 and concurrently resulting in the mean logits of \textcolor{OliveGreen}{gold answer token} surpassing the \textcolor{Red}{initial wrong answer in Italian}; \textbf{Later layers (25-32): }Patching $h_{\text{Eng}}^{\lambda}$ in the later layers at the penultimate token is not very effective since the Italian forward pass has already executed the decision and resists changing its original incorrect prediction (indicated by the steep drop in \% flip in the later layers). This behavior contrasts with patching the last token (\autoref{fig:belebele_lasttokenpatch_main}), where the \% flip steadily increases as we patch in the later layers, since the last-token representations are directly associated with the next-token prediction.

Overall, for both token positions, the patched $h_{\text{Eng}}^{\lambda}$ in the middle layers resulted in higher mean logits associated with $y_{c}$, suggesting that transfer failures stem from the non-English forward pass constructing an \emph{`incorrect'} representation (different from $h_{\text{Eng}}^{\lambda}$) in these layers. While patching $h_{\text{Eng}}^{\lambda}$ does not flip all samples,  the mean logits of the gold token surpass the initial wrong answer on average. This suggests that misalignment in the middle layers is primarily responsible for incorrect predictions for languages other than English.

\subsection{What does $h_{\text{Eng}}^{\lambda}$ encode?}

\label{sec:control_results}

\begin{figure}[h]
    \centering
\includegraphics[width=0.86\linewidth]{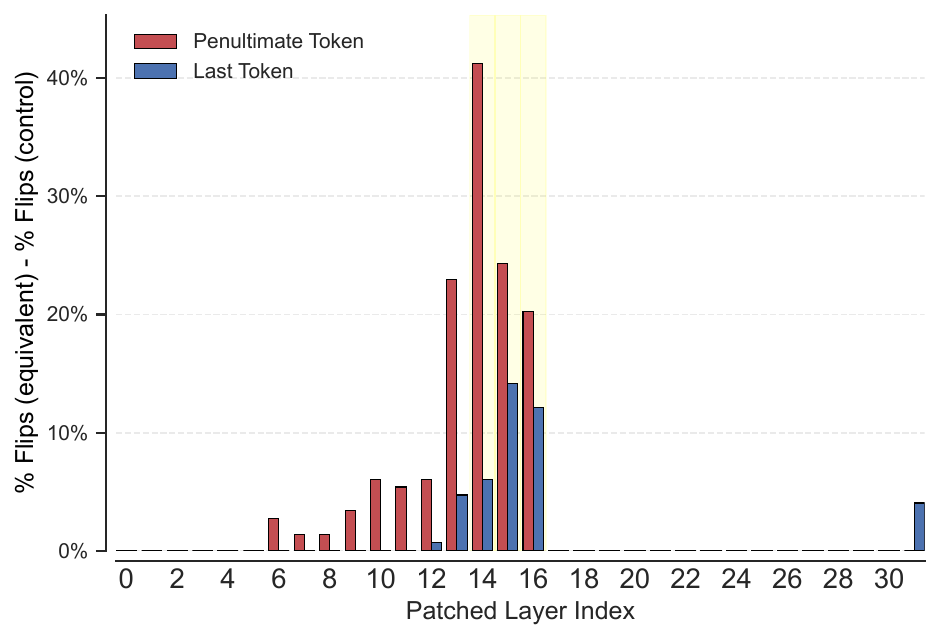}
    \caption{Control patching results - belebele Italian TF samples.}
\label{fig:belebele_control_patching_main_results}
\end{figure}

\autoref{fig:belebele_control_patching_main_results} shows the $\Delta$ \% flip for Italian TF samples on Belebele, measuring the \% of samples that flip exclusively under semantically matched patching relative to controls. We identify a select few middle layers where the $\Delta $(\% Flip) is high, indicating that the cross-lingual semantic information is causally involved in driving the model’s prediction. For example, in layer 14 of the penultimate token patching (\autoref{fig:belebele_control_patching_main_results}), 41.2\% (out of 68.9\% total flips) can be flipped only by patching semantically equivalent English sample activation. We see this behavior consistent in other languages as well (\autoref{fig:control_exp_results}). While this confirms that cross-lingual conceptual alignment in the middle layers mediates correctness, it calls for future work to disentangle the effects of conceptual alignment from those of next-token production in the representations.
\section{Discussion}
Beyond our core analysis, we explore several secondary results that clarify the nature of cross-lingual representations. This includes an instance-level qualitative analysis ($\S$\ref{sec:Qualitativeanalysis}), an investigation into language-specific trends ($\S$\ref{sec:langspecifictrend}), and an assessment of model entropy ($\S$\ref{sec:entropy}).
\subsection{Qualitative analysis}
\label{sec:Qualitativeanalysis}
We illustrate a false positive instance misclassified by $\DALI$ from the XStorycloze Hindi benchmark. At layer 14, 92.9\% of the samples have $\DALI$ = 1, but only 64.9\% have $\DALIStrong$ = 1, indicating false positives of alignment in $\DALI$, but corrected by the intra-lingual constraint in $\DALIStrong$.

\begin{figure}[h!]
    \centering
\includegraphics[width=\linewidth]{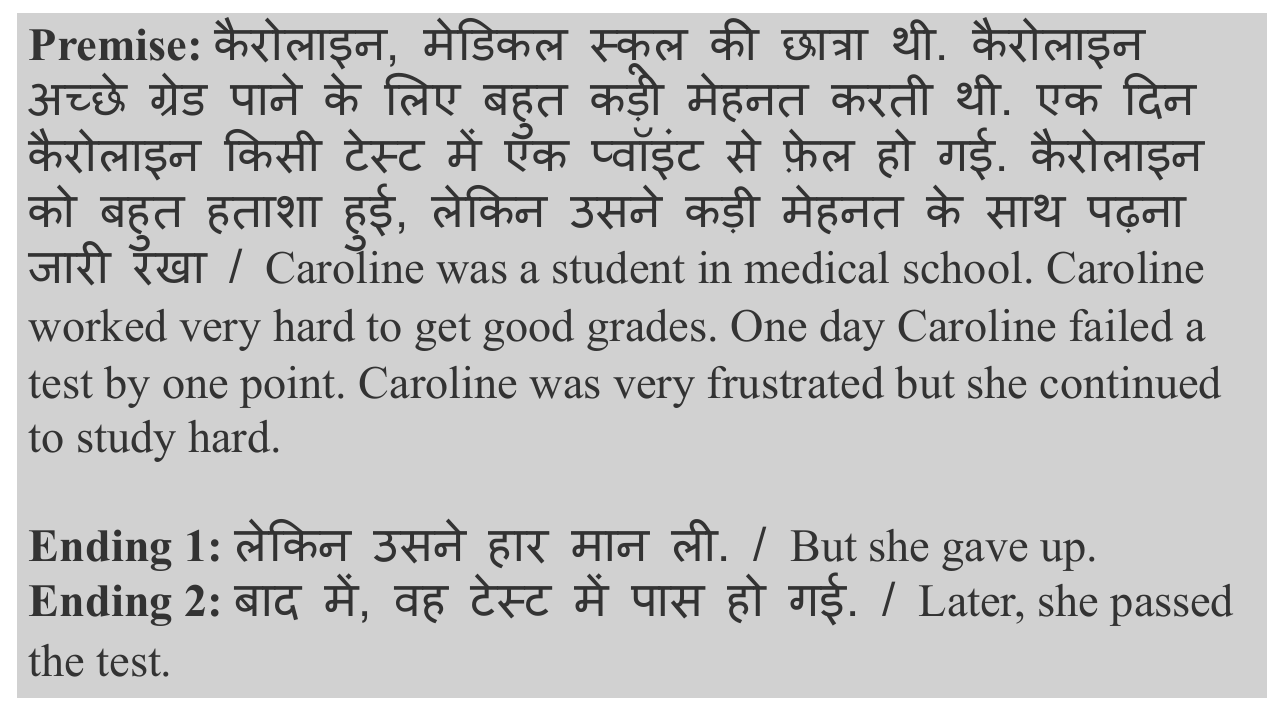}
        \caption{TF Hindi instance in XStorycloze.}
        \label{fig:qualitative_ex}
\end{figure}

One such example ($\DALI=1; \DALIStrong=0$) is shown in \autoref{fig:qualitative_ex}. Patching experiments reveal that the instance flips only when patching the last-token activation at layer 30. No correction is observed at earlier layers, indicating that the Hindi forward pass remains corrupted and is only corrected by a next-token decision signal, rather than by semantic alignment. This sample was classified as ‘aligned’ by $\DALI$, but \emph{caught} by $\DALIStrong$,  illustrating how the proposed alignment metrics differ. 
\subsection{Language-specific trends of alignment}
\label{sec:langspecifictrend}
We analyzed the \% of samples with alignment = 1 ($\DALI, \DALIStrong, \MEXAtask$) across the LLM layers (\autoref{fig:alignment_trajectories}). We found a consistent pattern: languages that are typologically closer to English and HR languages (e.g., Spanish, German) exhibit higher alignment than languages that are not (e.g., Hindi, Thai). We also found that cross-lingual alignment peaks in the middle layers, consistent with prior works \citep{kargaran-etal-2025-mexa, wilie-etal-2025-high, liu-niehues-2025-middle}.
\subsection{Entropy of equivalent and control patches}
\label{sec:entropy}
Both semantically matched, and control patches induce prediction flips in the TF examples (\autoref{fig:belebele_control_patching_main_results} - only some layers have non-zero $\Delta$s). A \emph{flip} to a correct answer in a discriminative instance is binary, and we investigate this further from the perspective of the entropy of the model's decision. 
\begin{equation}
    \text{entropy} = -\sum_{t\in T}p_t \times log(p_t)
\end{equation}

\begin{figure}[h]
    \centering
\includegraphics[width=0.9\linewidth]{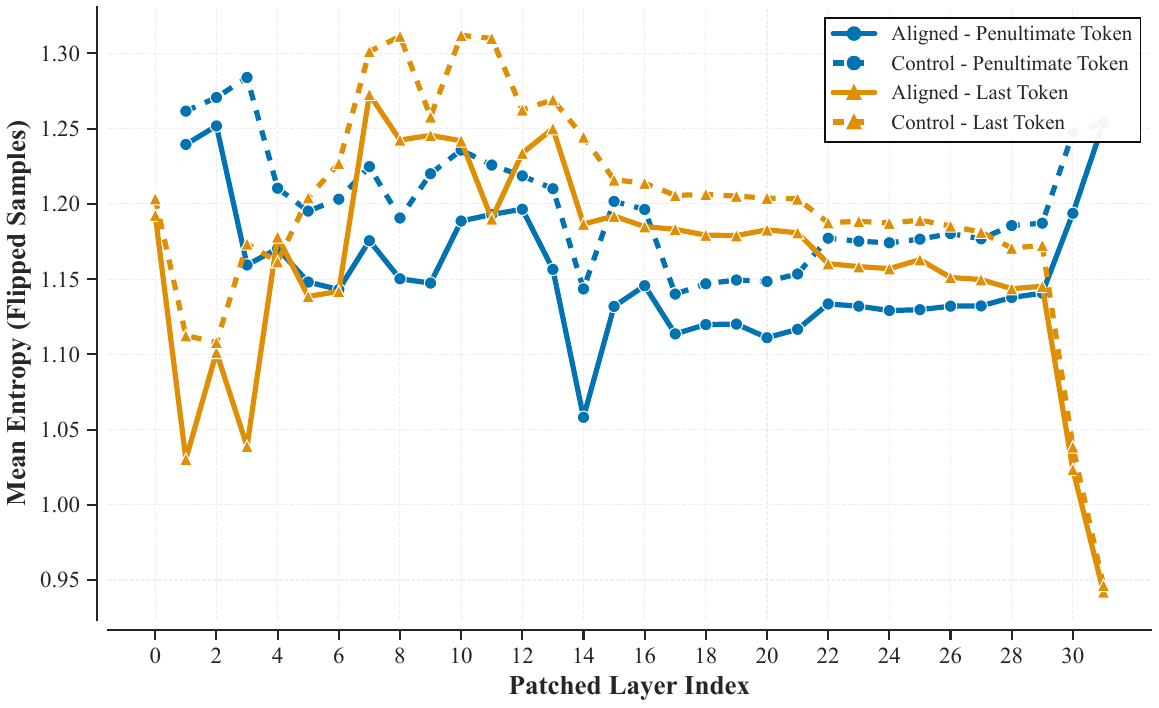}
        \caption{Entropy analysis of belebele French samples.}
\label{fig:entropy_analysis}
\end{figure}

We observe (\autoref{fig:entropy_analysis}) that the flips triggered by semantically equivalent English patches are consistently associated with lower output entropy. This suggests that semantic alignment not only increases the likelihood of correction but also stabilizes the model’s decision, yielding more confident predictions than control patches that primarily inject a next-token bias. This analysis highlights an underexplored area of the effect of cross-lingual alignment on confidence calibration. 
\section{Related Work}
\paragraph{Multilingual LLMs}
Multilingual LLMs are designed to process and generate text across multiple languages. However, the pretraining corpus of most state-of-the-art LLMs is dominated by English, despite exhibiting reasonable capabilities in other languages \citep{ahia-etal-2023-languages}.  \citet{etxaniz2023multilinguallanguagemodelsthink} provided evidence that multilingual LLMs perform better in English through \textit{`self-translate'}, where LLMs were first instructed to translate the other-language prompts to English and process them in English. \citet{wendler-etal-2024-llamas} extended this by decoding middle layer residuals to show that LLMs place more probability mass on English tokens than on tokens of the prompted language before transitioning to the target-language vocabulary in the final layers -- a pattern which is interpreted as reasoning in an abstract, language-agnostic concept space in the middle layers biased toward English rather than reliance on English as an explicit lexical pivot. The presence of a language-agnostic concept space was validated by \citet{dumas-etal-2025-separating} using activation patching. Multiple concurrent studies provided evidence for an implicit translation $\to$ task-solving $\to$ translation pipeline underlying improved performance in languages other than English, positing through early decoding that models \emph{`solve'} the task by decoding the right English token in the middle layers but often fail to faithfully translate the resulting \emph{`gold'} token back into the target language during final decoding \citep{bafna2025translationbarrierhypothesismultilingual, lu-etal-2025-paths, wang-etal-2025-lost-multilinguality}. 
\citet{10.1609/aaai.v39i27.35038} and \citet{kargaran-etal-2025-mexa} demonstrated that the representational similarity between parallel non-English and English corpora, independent from the task, acts as a good barometer for multilingual performance. Our study extends the concept of representational alignment to the instance level and provides causal evidence that middle-layer alignment is sufficient for the reasoning required in MCQA tasks.

\paragraph{Boosting cross-lingual alignment}
Recent work has sought to enhance alignment through interventions, either by incorporating a separate cross-lingual alignment objective \citep{liu-niehues-2025-middle} or fine-tuning \citep{li2024improvingincontextlearningmultilingual,zhang2023baylingbridgingcrosslingualalignment}, and has shown that improved alignment yields gains in downstream multilingual accuracy. Concurrent works also analyze inference time steering to improve multilingual performance \citep{wang-etal-2025-bridging, lim2025languagespecificlatentprocesshinders,sundar-etal-2025-steering}. The idea is to steer the other language representations towards the shared latent space by computing a steering vector or a projection matrix. Our study is similar, but instead of steering representations effectively, we demonstrate how \emph{`forcing'} alignment by patching the English activation in the middle layers helps the model rectify non-English failures.

\section{Conclusions}
This paper introduces instance-level alignment metrics to study LLM performance disparities across languages in discriminative tasks. We demonstrate that samples that generalize well between English and languages other than English (TS) exhibit a higher degree of alignment than samples that do not (TF), thereby establishing that cross-lingual alignment with English at the instance level is correlated with task success. To probe the causal role of alignment, we perform activation patching experiments on non-English failure instances. While patching English activations can correct incorrect non-English predictions, our control experiments help us identify specific middle layers that flip \emph{only} with semantically equivalent patches. Overall, these findings suggest that cross-lingual semantic alignment in localized middle layers contributes causally to non-English correctness. These findings motivate future directions such as exploring targeted steering at inference time that leverages the model's high-fidelity English activations to enhance performance on languages other than English, while also studying potential downsides, including inflated confidence and the erasure of valuable local, non-English knowledge.
\section*{Limitations}
Firstly, the study's limitations center on its scope, which is restricted to analyzing discriminative NLU tasks. This is due to the necessity of multilingual benchmarks that are parallel across languages. Secondly, as with most multilingual benchmarks, the benchmarks considered in the study were initially constructed in English and translated into other languages by humans, which could introduce translation artifacts \citep{artetxe-etal-2020-translation}. Thirdly, our study analyzes only the bilingual alignment of languages other than English with English, not between other languages. 

\section*{Acknowledgements}
We would like to thank the members of the CLIP
lab at the University of Maryland for their constructive feedback and support. Specifically, we are grateful to Navita Goyal, Hiba El Oirghi, Dayeon (Zoey) Ki, Calvin Bao, Eric Bennet, Hieu Tran, and Osvaldo Quinjica for extensive discussions on the analyses. The authors also acknowledge the assistance of LLMs in developing this paper, in particular, using AI agents such as Claude Code to write code and create plots. 

\clearpage
\bibliography{anthology,custom}

@misc{touvron2023llama2openfoundation,
      title={Llama 2: Open Foundation and Fine-Tuned Chat Models}, 
      author={Hugo Touvron and Louis Martin and Kevin Stone and Peter Albert and Amjad Almahairi and Yasmine Babaei and Nikolay Bashlykov and Soumya Batra and Prajjwal Bhargava and Shruti Bhosale and others},
      year={2023},
      eprint={2307.09288},
      archivePrefix={arXiv},
      primaryClass={cs.CL},
      url={https://arxiv.org/abs/2307.09288}, 
}

@inproceedings{muennighoff-etal-2023-crosslingual,
    title = "Crosslingual Generalization through Multitask Finetuning",
    author = "Muennighoff, Niklas  and
      Wang, Thomas  and
      Sutawika, Lintang  and
      Roberts, Adam  and
      Biderman, Stella  and
      Le Scao, Teven  and
      Bari, M Saiful  and
      Shen, Sheng  and
      Yong, Zheng Xin  and
      Schoelkopf, Hailey  and
      others",
    editor = "Rogers, Anna  and
      Boyd-Graber, Jordan  and
      Okazaki, Naoaki",
    booktitle = "Proceedings of the 61st Annual Meeting of the Association for Computational Linguistics (Volume 1: Long Papers)",
    month = jul,
    year = "2023",
    address = "Toronto, Canada",
    publisher = "Association for Computational Linguistics",
    url = "https://aclanthology.org/2023.acl-long.891/",
    doi = "10.18653/v1/2023.acl-long.891",
    pages = "15991--16111"
}

@inproceedings{hammerl-etal-2024-understanding,
    title = "Understanding Cross-Lingual {A}lignment{---}{A} Survey",
    author = {H{\"a}mmerl, Katharina  and
      Libovick{\'y}, Jind{\v{r}}ich  and
      Fraser, Alexander},
    editor = "Ku, Lun-Wei  and
      Martins, Andre  and
      Srikumar, Vivek",
    booktitle = "Findings of the Association for Computational Linguistics: ACL 2024",
    month = aug,
    year = "2024",
    address = "Bangkok, Thailand",
    publisher = "Association for Computational Linguistics",
    url = "https://aclanthology.org/2024.findings-acl.649/",
    doi = "10.18653/v1/2024.findings-acl.649",
    pages = "10922--10943"
}

@inproceedings{wendler-etal-2024-llamas,
    title = "Do Llamas Work in {E}nglish? On the Latent Language of Multilingual Transformers",
    author = "Wendler, Chris  and
      Veselovsky, Veniamin  and
      Monea, Giovanni  and
      West, Robert",
    editor = "Ku, Lun-Wei  and
      Martins, Andre  and
      Srikumar, Vivek",
    booktitle = "Proceedings of the 62nd Annual Meeting of the Association for Computational Linguistics (Volume 1: Long Papers)",
    month = aug,
    year = "2024",
    address = "Bangkok, Thailand",
    publisher = "Association for Computational Linguistics",
    url = "https://aclanthology.org/2024.acl-long.820/",
    doi = "10.18653/v1/2024.acl-long.820",
    pages = "15366--15394"
}

@inproceedings{kargaran-etal-2025-mexa,
    title = "{MEXA}: Multilingual Evaluation of {E}nglish-Centric {LLM}s via Cross-Lingual Alignment",
    author = "Kargaran, Amir Hossein  and
      Modarressi, Ali  and
      Nikeghbal, Nafiseh  and
      Diesner, Jana  and
      Yvon, Fran{\c{c}}ois  and
      Schuetze, Hinrich",
    editor = "Che, Wanxiang  and
      Nabende, Joyce  and
      Shutova, Ekaterina  and
      Pilehvar, Mohammad Taher",
    booktitle = "Findings of the Association for Computational Linguistics: ACL 2025",
    month = jul,
    year = "2025",
    address = "Vienna, Austria",
    publisher = "Association for Computational Linguistics",
    url = "https://aclanthology.org/2025.findings-acl.1385/",
    doi = "10.18653/v1/2025.findings-acl.1385",
    pages = "27001--27023",
    ISBN = "979-8-89176-256-5"
}

@misc{nllbteam2022languageleftbehindscaling,
      title={No Language Left Behind: Scaling Human-Centered Machine Translation}, 
      author={NLLB Team and Marta R. Costa-jussà and James Cross and Onur Çelebi and Maha Elbayad and Kenneth Heafield and Kevin Heffernan and Elahe Kalbassi and Janice Lam and Daniel Licht and Jean Maillard and others},
      year={2022},
      eprint={2207.04672},
      archivePrefix={arXiv},
      primaryClass={cs.CL},
      url={https://arxiv.org/abs/2207.04672}, 
}

@inproceedings{bandarkar-etal-2024-belebele,
    title = "The Belebele Benchmark: a Parallel Reading Comprehension Dataset in 122 Language Variants",
    author = "Bandarkar, Lucas  and
      Liang, Davis  and
      Muller, Benjamin  and
      Artetxe, Mikel  and
      Shukla, Satya Narayan  and
      Husa, Donald  and
      Goyal, Naman  and
      Krishnan, Abhinandan  and
      Zettlemoyer, Luke  and
      Khabsa, Madian",
    editor = "Ku, Lun-Wei  and
      Martins, Andre  and
      Srikumar, Vivek",
    booktitle = "Proceedings of the 62nd Annual Meeting of the Association for Computational Linguistics (Volume 1: Long Papers)",
    month = aug,
    year = "2024",
    address = "Bangkok, Thailand",
    publisher = "Association for Computational Linguistics",
    url = "https://aclanthology.org/2024.acl-long.44/",
    doi = "10.18653/v1/2024.acl-long.44",
    pages = "749--775"
}

@inproceedings{lin-etal-2022-shot,
    title = "Few-shot Learning with Multilingual Generative Language Models",
    author = "Lin, Xi Victoria  and
      Mihaylov, Todor  and
      Artetxe, Mikel  and
      Wang, Tianlu  and
      Chen, Shuohui  and
      Simig, Daniel  and
      Ott, Myle  and
      Goyal, Naman  and
      Bhosale, Shruti  and
      Du, Jingfei  and
      others",
    editor = "Goldberg, Yoav  and
      Kozareva, Zornitsa  and
      Zhang, Yue",
    booktitle = "Proceedings of the 2022 Conference on Empirical Methods in Natural Language Processing",
    month = dec,
    year = "2022",
    address = "Abu Dhabi, United Arab Emirates",
    publisher = "Association for Computational Linguistics",
    url = "https://aclanthology.org/2022.emnlp-main.616/",
    doi = "10.18653/v1/2022.emnlp-main.616",
    pages = "9019--9052"
}

@inproceedings{vig2020investigating,
  title={Investigating Gender Bias in Language Models Using Causal Mediation Analysis},
  author={Vig, Jesse and Gehrmann, Sebastian and Belinkov, Yonatan and Qian, Sharon and Nevo, Daniel and Singer, Yaron and Shieber, Stuart},
  booktitle={Advances in Neural Information Processing Systems},
  volume={33},
  pages={12388--12401},
  year={2020}
}

@inproceedings{10.1609/aaai.v39i27.35038,
author = {Li, Zihao and Shi, Yucheng and Liu, Zirui and Yang, Fan and Payani, Ali and Liu, Ninghao and Du, Mengnan},
title = {Language ranker: a metric for quantifying LLM performance across high and low-resource languages},
year = {2025},
isbn = {978-1-57735-897-8},
publisher = {AAAI Press},
url = {https://doi.org/10.1609/aaai.v39i27.35038},
doi = {10.1609/aaai.v39i27.35038},
booktitle = {Proceedings of the Thirty-Ninth AAAI Conference on Artificial Intelligence and Thirty-Seventh Conference on Innovative Applications of Artificial Intelligence and Fifteenth Symposium on Educational Advances in Artificial Intelligence},
articleno = {3140},
numpages = {9},
series = {AAAI'25/IAAI'25/EAAI'25}
}

@inproceedings{dumas-etal-2025-separating,
    title = "Separating Tongue from Thought: Activation Patching Reveals Language-Agnostic Concept Representations in Transformers",
    author = "Dumas, Cl{\'e}ment  and
      Wendler, Chris  and
      Veselovsky, Veniamin  and
      Monea, Giovanni  and
      West, Robert",
    editor = "Che, Wanxiang  and
      Nabende, Joyce  and
      Shutova, Ekaterina  and
      Pilehvar, Mohammad Taher",
    booktitle = "Proceedings of the 63rd Annual Meeting of the Association for Computational Linguistics (Volume 1: Long Papers)",
    month = jul,
    year = "2025",
    address = "Vienna, Austria",
    publisher = "Association for Computational Linguistics",
    url = "https://aclanthology.org/2025.acl-long.1536/",
    doi = "10.18653/v1/2025.acl-long.1536",
    pages = "31822--31841",
    ISBN = "979-8-89176-251-0"
}

@misc{neelakantan2022textcodeembeddingscontrastive,
      title={Text and Code Embeddings by Contrastive Pre-Training}, 
      author={Arvind Neelakantan and Tao Xu and Raul Puri and Alec Radford and Jesse Michael Han and Jerry Tworek and Qiming Yuan and Nikolas Tezak and Jong Wook Kim and Chris Hallacy and others},
      year={2022},
      eprint={2201.10005},
      archivePrefix={arXiv},
      primaryClass={cs.CL},
      url={https://arxiv.org/abs/2201.10005}, 
}

@inproceedings{wang-etal-2024-improving-text,
    title = "Improving Text Embeddings with Large Language Models",
    author = "Wang, Liang  and
      Yang, Nan  and
      Huang, Xiaolong  and
      Yang, Linjun  and
      Majumder, Rangan  and
      Wei, Furu",
    editor = "Ku, Lun-Wei  and
      Martins, Andre  and
      Srikumar, Vivek",
    booktitle = "Proceedings of the 62nd Annual Meeting of the Association for Computational Linguistics (Volume 1: Long Papers)",
    month = aug,
    year = "2024",
    address = "Bangkok, Thailand",
    publisher = "Association for Computational Linguistics",
    url = "https://aclanthology.org/2024.acl-long.642/",
    doi = "10.18653/v1/2024.acl-long.642",
    pages = "11897--11916"
}

@inproceedings{ahia-etal-2023-languages,
    title = "Do All Languages Cost the Same? Tokenization in the Era of Commercial Language Models",
    author = "Ahia, Orevaoghene  and
      Kumar, Sachin  and
      Gonen, Hila  and
      Kasai, Jungo  and
      Mortensen, David  and
      Smith, Noah  and
      Tsvetkov, Yulia",
    editor = "Bouamor, Houda  and
      Pino, Juan  and
      Bali, Kalika",
    booktitle = "Proceedings of the 2023 Conference on Empirical Methods in Natural Language Processing",
    month = dec,
    year = "2023",
    address = "Singapore",
    publisher = "Association for Computational Linguistics",
    url = "https://aclanthology.org/2023.emnlp-main.614/",
    doi = "10.18653/v1/2023.emnlp-main.614",
    pages = "9904--9923"
}

@misc{etxaniz2023multilinguallanguagemodelsthink,
      title={Do Multilingual Language Models Think Better in English?}, 
      author={Julen Etxaniz and Gorka Azkune and Aitor Soroa and Oier Lopez de Lacalle and Mikel Artetxe},
      year={2023},
      eprint={2308.01223},
      archivePrefix={arXiv},
      primaryClass={cs.CL},
      url={https://arxiv.org/abs/2308.01223}, 
}

@misc{zhang2023baylingbridgingcrosslingualalignment,
      title={BayLing: Bridging Cross-lingual Alignment and Instruction Following through Interactive Translation for Large Language Models}, 
      author={Shaolei Zhang and Qingkai Fang and Zhuocheng Zhang and Zhengrui Ma and Yan Zhou and Langlin Huang and Mengyu Bu and Shangtong Gui and Yunji Chen and Xilin Chen and Yang Feng},
      year={2023},
      eprint={2306.10968},
      archivePrefix={arXiv},
      primaryClass={cs.CL},
      url={https://arxiv.org/abs/2306.10968}, 
}

@inproceedings{lu-etal-2025-paths,
    title = "Paths Not Taken: Understanding and Mending the Multilingual Factual Recall Pipeline",
    author = "Lu, Meng  and
      Zhang, Ruochen  and
      Eickhoff, Carsten  and
      Pavlick, Ellie",
    editor = "Christodoulopoulos, Christos  and
      Chakraborty, Tanmoy  and
      Rose, Carolyn  and
      Peng, Violet",
    booktitle = "Proceedings of the 2025 Conference on Empirical Methods in Natural Language Processing",
    month = nov,
    year = "2025",
    address = "Suzhou, China",
    publisher = "Association for Computational Linguistics",
    url = "https://aclanthology.org/2025.emnlp-main.762/",
    doi = "10.18653/v1/2025.emnlp-main.762",
    pages = "15077--15107",
    ISBN = "979-8-89176-332-6"
}

@inproceedings{wang-etal-2025-lost-multilinguality,
    title = "Lost in Multilinguality: Dissecting Cross-lingual Factual Inconsistency in Transformer Language Models",
    author = {Wang, Mingyang  and
      Adel, Heike  and
      Lange, Lukas  and
      Liu, Yihong  and
      Nie, Ercong  and
      Str{\"o}tgen, Jannik  and
      Schuetze, Hinrich},
    editor = "Che, Wanxiang  and
      Nabende, Joyce  and
      Shutova, Ekaterina  and
      Pilehvar, Mohammad Taher",
    booktitle = "Proceedings of the 63rd Annual Meeting of the Association for Computational Linguistics (Volume 1: Long Papers)",
    month = jul,
    year = "2025",
    address = "Vienna, Austria",
    publisher = "Association for Computational Linguistics",
    url = "https://aclanthology.org/2025.acl-long.253/",
    doi = "10.18653/v1/2025.acl-long.253",
    pages = "5075--5094",
    ISBN = "979-8-89176-251-0"
}

@misc{lim2025languagespecificlatentprocesshinders,
      title={Language-Specific Latent Process Hinders Cross-Lingual Performance}, 
      author={Zheng Wei Lim and Alham Fikri Aji and Trevor Cohn},
      year={2025},
      eprint={2505.13141},
      archivePrefix={arXiv},
      primaryClass={cs.CL},
      url={https://arxiv.org/abs/2505.13141}, 
}

@misc{li2024improvingincontextlearningmultilingual,
      title={Improving In-context Learning of Multilingual Generative Language Models with Cross-lingual Alignment}, 
      author={Chong Li and Shaonan Wang and Jiajun Zhang and Chengqing Zong},
      year={2024},
      eprint={2311.08089},
      archivePrefix={arXiv},
      primaryClass={cs.CL},
      url={https://arxiv.org/abs/2311.08089}, 
}

@inproceedings{wang-etal-2025-bridging,
    title = "Bridging the Language Gaps in Large Language Models with Inference-Time Cross-Lingual Intervention",
    author = "Wang, Weixuan  and
      Wu, Minghao  and
      Haddow, Barry  and
      Birch, Alexandra",
    editor = "Che, Wanxiang  and
      Nabende, Joyce  and
      Shutova, Ekaterina  and
      Pilehvar, Mohammad Taher",
    booktitle = "Proceedings of the 63rd Annual Meeting of the Association for Computational Linguistics (Volume 1: Long Papers)",
    month = jul,
    year = "2025",
    address = "Vienna, Austria",
    publisher = "Association for Computational Linguistics",
    url = "https://aclanthology.org/2025.acl-long.270/",
    doi = "10.18653/v1/2025.acl-long.270",
    pages = "5418--5433",
    ISBN = "979-8-89176-251-0"
}

@misc{grattafiori2024llama3herdmodels,
      title={The Llama 3 Herd of Models}, 
      author={Aaron Grattafiori and Abhimanyu Dubey and Abhinav Jauhri and Abhinav Pandey and Abhishek Kadian and Ahmad Al-Dahle and Aiesha Letman and Akhil Mathur and Alan Schelten and Alex Vaughan and others},
      year={2024},
      eprint={2407.21783},
      archivePrefix={arXiv},
      primaryClass={cs.AI},
      url={https://arxiv.org/abs/2407.21783}, 
}

@misc{nostalgebraist, title={Interpreting gpt: The logit lens; Accessed: 2025-09-25},
author={nosalgebraist},
year={2020},
url={https://lesswrong.com/posts/AcKRB8wDpdaN6v6ru}
}

@misc{bafna2025translationbarrierhypothesismultilingual,
      title={The Translation Barrier Hypothesis: Multilingual Generation with Large Language Models Suffers from Implicit Translation Failure}, 
      author={Niyati Bafna and Tianjian Li and Kenton Murray and David R. Mortensen and David Yarowsky and Hale Sirin and Daniel Khashabi},
      year={2025},
      eprint={2506.22724},
      archivePrefix={arXiv},
      primaryClass={cs.CL},
      url={https://arxiv.org/abs/2506.22724}, 
}

@inproceedings{liu-niehues-2025-middle,
    title = "Middle-Layer Representation Alignment for Cross-Lingual Transfer in Fine-Tuned {LLM}s",
    author = "Liu, Danni  and
      Niehues, Jan",
    editor = "Che, Wanxiang  and
      Nabende, Joyce  and
      Shutova, Ekaterina  and
      Pilehvar, Mohammad Taher",
    booktitle = "Proceedings of the 63rd Annual Meeting of the Association for Computational Linguistics (Volume 1: Long Papers)",
    month = jul,
    year = "2025",
    address = "Vienna, Austria",
    publisher = "Association for Computational Linguistics",
    url = "https://aclanthology.org/2025.acl-long.778/",
    doi = "10.18653/v1/2025.acl-long.778",
    pages = "15979--15996",
    ISBN = "979-8-89176-251-0"
}

@article{DBLP:journals/corr/abs-2408-01416,
  publtype={informal},
  author={Aaron Mueller and Jannik Brinkmann and Millicent L. Li and Samuel Marks and Koyena Pal and Nikhil Prakash and Can Rager and Aruna Sankaranarayanan and Arnab Sen Sharma and Jiuding Sun and Eric Todd and David Bau and Yonatan Belinkov},
  title={The Quest for the Right Mediator: A History, Survey, and Theoretical Grounding of Causal Interpretability},
  year={2024},
  cdate={1704067200000},
  journal={CoRR},
  volume={abs/2408.01416},
  url={https://doi.org/10.48550/arXiv.2408.01416}
}

@misc{fiottokaufman2025nnsightndifdemocratizingaccess,
      title={NNsight and NDIF: Democratizing Access to Open-Weight Foundation Model Internals}, 
      author={Jaden Fiotto-Kaufman and Alexander R. Loftus and Eric Todd and Jannik Brinkmann and Koyena Pal and Dmitrii Troitskii and Michael Ripa and Adam Belfki and Can Rager and Caden Juang and others},
      year={2025},
      eprint={2407.14561},
      archivePrefix={arXiv},
      primaryClass={cs.LG},
      url={https://arxiv.org/abs/2407.14561}, 
}

@inproceedings{
wiegreffe2025answerassembleaceunderstanding,
title={Answer, Assemble, Ace: Understanding How {LM}s Answer Multiple Choice Questions},
author={Sarah Wiegreffe and Oyvind Tafjord and Yonatan Belinkov and Hannaneh Hajishirzi and Ashish Sabharwal},
booktitle={The Thirteenth International Conference on Learning Representations},
year={2025},
url={https://openreview.net/forum?id=6NNA0MxhCH}
}

@inproceedings{
meng2022locating,
title={Locating afnd Editing Factual Associations in {GPT}},
author={Kevin Meng and David Bau and Alex J Andonian and Yonatan Belinkov},
booktitle={Advances in Neural Information Processing Systems},
editor={Alice H. Oh and Alekh Agarwal and Danielle Belgrave and Kyunghyun Cho},
year={2022},
url={https://openreview.net/forum?id=-h6WAS6eE4}
}

@misc{aryabumi2024aya23openweight,
      title={Aya 23: Open Weight Releases to Further Multilingual Progress}, 
      author={Viraat Aryabumi and John Dang and Dwarak Talupuru and Saurabh Dash and David Cairuz and Hangyu Lin and Bharat Venkitesh and Madeline Smith and Jon Ander Campos and Yi Chern Tan and others},
      year={2024},
      eprint={2405.15032},
      archivePrefix={arXiv},
      primaryClass={cs.CL},
      url={https://arxiv.org/abs/2405.15032}, 
}

@inproceedings{sundar-etal-2025-steering,
    title = "Steering into New Embedding Spaces: Analyzing Cross-Lingual Alignment Induced by Model Interventions in Multilingual Language Models",
    author = "Sundar, Anirudh  and
      Williamson, Sinead  and
      Metcalf, Katherine  and
      Theobald, Barry-John  and
      Seto, Skyler  and
      Fedzechkina, Masha",
    editor = "Che, Wanxiang  and
      Nabende, Joyce  and
      Shutova, Ekaterina  and
      Pilehvar, Mohammad Taher",
    booktitle = "Proceedings of the 63rd Annual Meeting of the Association for Computational Linguistics (Volume 1: Long Papers)",
    month = jul,
    year = "2025",
    address = "Vienna, Austria",
    publisher = "Association for Computational Linguistics",
    url = "https://aclanthology.org/2025.acl-long.118/",
    doi = "10.18653/v1/2025.acl-long.118",
    pages = "2375--2401",
    ISBN = "979-8-89176-251-0"
}

@inproceedings{wilie-etal-2025-high,
    title = "High-Dimensional Interlingual Representations of Large Language Models",
    author = "Wilie, Bryan  and
      Cahyawijaya, Samuel  and
      He, Junxian  and
      Fung, Pascale",
    editor = "Hahn, Michael  and
      Rani, Priya  and
      Kumar, Ritesh  and
      Shcherbakov, Andreas  and
      Sorokin, Alexey  and
      Serikov, Oleg  and
      Cotterell, Ryan  and
      Vylomova, Ekaterina",
    booktitle = "Proceedings of the 7th Workshop on Research in Computational Linguistic Typology and Multilingual NLP",
    month = aug,
    year = "2025",
    address = "Vienna, Austria",
    publisher = "Association for Computational Linguistics",
    url = "https://aclanthology.org/2025.sigtyp-1.14/",
    doi = "10.18653/v1/2025.sigtyp-1.14",
    pages = "122--155",
    ISBN = "979-8-89176-281-7"
}
\bibliographystyle{acl_natbib}
\clearpage

\appendix

\section{Appendix}
 \begin{figure}[h]
    \centering
    \begin{subfigure}[b]{\columnwidth}
        \centering
        \includegraphics[width=\columnwidth]{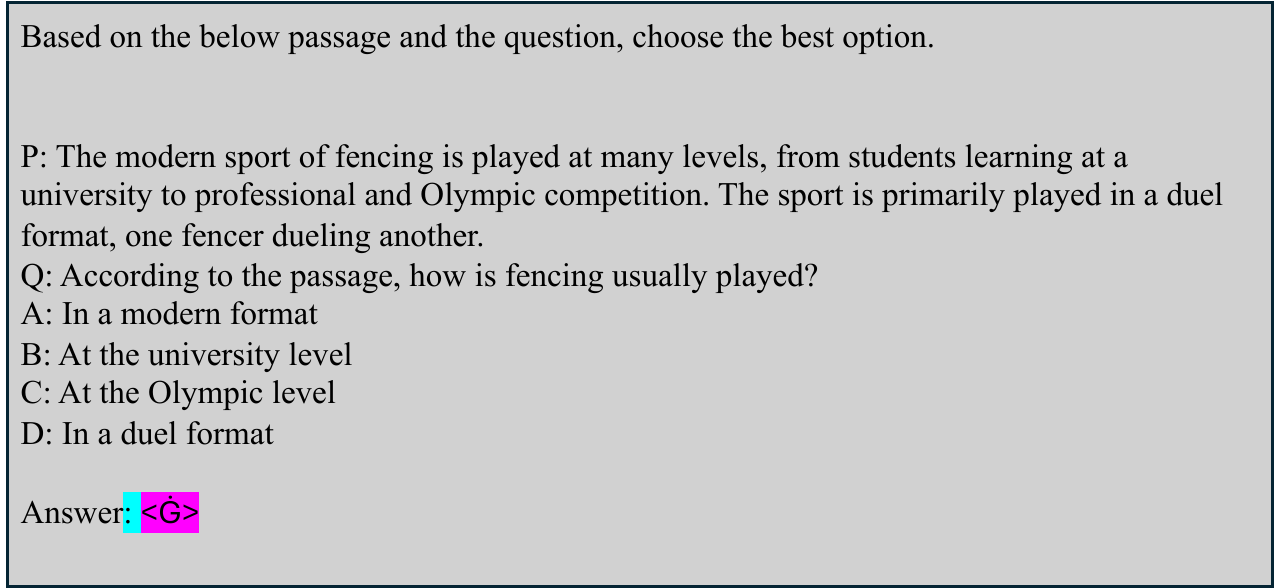}
        \caption{Belebele evaluation prompt}
        \label{fig:belebele_prompt}
    \end{subfigure}
    
    \vspace{1em}
    
    \begin{subfigure}[b]{\columnwidth}
        \centering
        \includegraphics[width=\columnwidth]{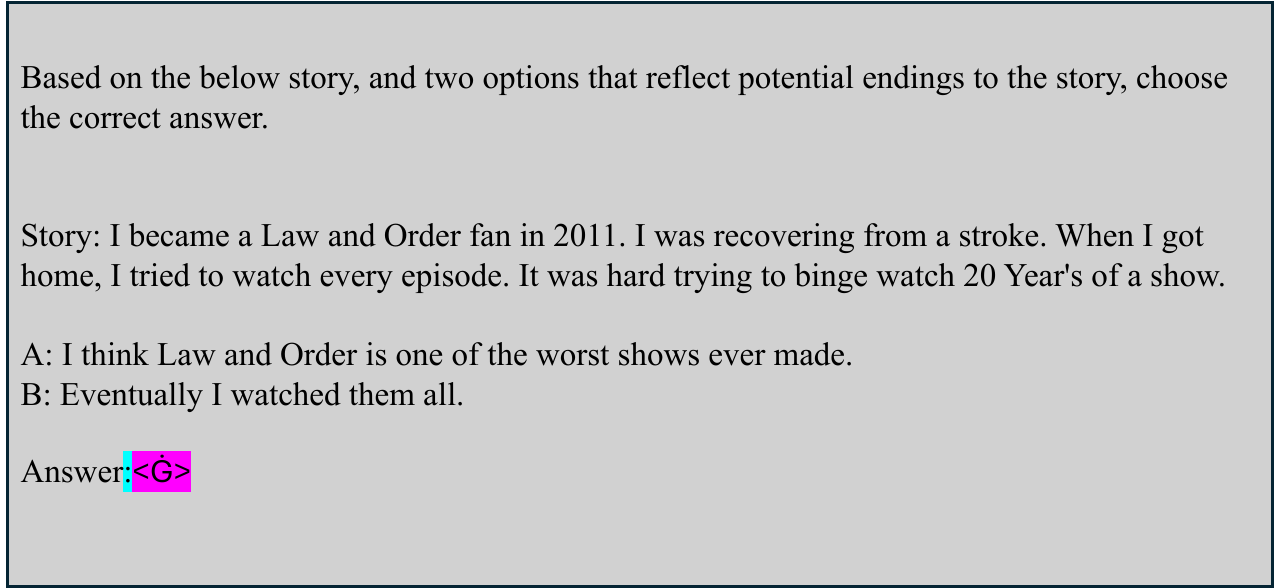}
        \caption{Xstorycloze evaluation prompt}
        \label{fig:xstorycloze_prompt}
    \end{subfigure}
    
    \vspace{1em}
    
    \begin{subfigure}[b]{\columnwidth}
        \centering
        \includegraphics[width=\columnwidth]{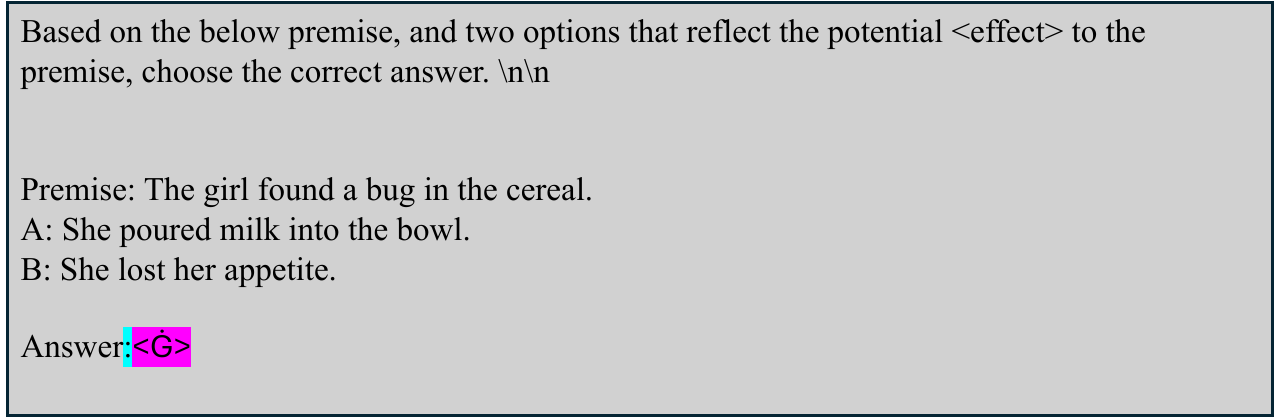}
        \caption{Xcopa evaluation prompt}
        \label{fig:xcopa_prompt}
    \end{subfigure}
    
    \caption{Evaluation prompts for different benchmarks. \colorbox{Aqua}{:} refers to the penultimate token and \colorbox{magenta}{Ġ} refers to the last token respectively.}
    \label{fig:evaluation_prompts}
\end{figure}
\clearpage

\begin{table*}[htbp] 
\small 
\begin{tabular}{p{1.9 cm}|p{6.0cm}p{3cm}p{3cm}}
\hline \textbf{Model} & \textbf{Belebele} & \textbf{Xstorycloze} & \textbf{Xcopa} \\ 
\hline 
Llama3.1 8B, Llama3.1 8B it & Spanish, Italian, French, German, Thai, Hindi, Portuguese & 
English, Spanish, Hindi &
Italian, Thai \\ 
\hline 
Aya23 8B & Modern Standard Arabic, Simplified Chinese, Traditional Chinese, Czech, Dutch, French, German, Greek, Hebrew, Hindi, Indonesian, Italian, Japanese, Korean, Persian, Polish, Portuguese, Romanian, Russian, Spanish, Turkish, Ukrainian, Vietnamese 
& Arabic, Spanish, Hindi, Indonesian, Russian, Chinese 
& Indonesian, Thai, Turkish, Vietnamese, Chinese \\ 
\hline 
\end{tabular} 
\caption{Models and languages used in the cross-lingual alignment experiments} 
\label{tab:models_lang} 
\end{table*}

\begin{table*}[!htbp]

\adjustbox{width=\textwidth}
{
\begin{tabular}{llccccccc}
\hline
 &  & \textbf{Spanish} & \textbf{Italian} & \textbf{French} & \textbf{German} & \textbf{Thai} & \textbf{Portuguese} & \textbf{Hindi} \\
\hline
\hline
\multirow{5}{*}{\textbf{Belebele (n=900)}} & ($n_{\text{TS}}$, $n_{\text{TF}}$) 
& (588, 135) & (575,148) & (608,115) & (586,137) & (482, 241) & (614, 109) & (466,257) \\
\cline{2-9}
& $\Delta\DALI (\text{p-val})$, $\lambda{\text{max}}$ & 7.5\% (0.06), 14& 5.9\% (0.10), 9 & 5.1\% (0.15), 14 & 6.8\% (0.08), 14 & 3.9\% (0.14), 14 & 0.3\% (0.48), 13 & 8.3\% (0.0), 14 \\
& $\Delta\DALIStrong (\text{p-val})$, $\lambda{\text{max}}$ & 2.6\% (0.28), 8 & 6.7\% (0.06), 9 & 6.6\% (0.08), 9 & 4.3\% (0.15), 8 & 1.9\% (0.15), 8 & -0.8\% (0.56), 9 & 0.4\% (0.4), 5 \\
& $\Delta\MEXAtask (\text{p-val})$, $\lambda_{\text{max}}$ & 6.5\% (0.08), 12 & 5\% (0.14), 12 & 11.9\% (0.01), 12 & 14.4\% (0), 7 & 2.5\% (0.09), 13 & 7.2\% (0.06), 12 & 6.8\% (0.02), 8 \\
\hline
\hline
\multirow{5}{*}{\textbf{Xstorycloze (n=1511)}} & ($n_{\text{TS}}$, $n_{\text{TF}}$) 
& (1365, 86) & . & . & . & . & . & (1240, 211) \\
\cline{2-9}
& $\Delta\DALI (\text{p-val})$, $\lambda_{\text{max}}$ & 3.8\% (0.01), 14 & . & . & . & . & . & 3.3\% (0.04), 14 \\
& $\Delta\DALIStrong (\text{p-val})$, $\lambda_{\text{max}}$ & 0.0\% (0.56), 13 & . & . & . & . & . & 9.5\% (0.0), 13 \\
& $\Delta\MEXAtask(\text{p-val})$, $\lambda_{\text{max}}$ & 0.0\% (0.60), 7${}^{*}$ & . & . & . & . & . & 2.0\% (0.01), 16 \\
\hline
\hline
\multirow{5}{*}{\textbf{Xcopa (n=500)}} & ($n_{\text{TS}}$, $n_{\text{TF}}$) & . & (374,155) & . & . & . & (262,94) & . \\
\cline{2-9}
& $\Delta\DALI (\text{p-val})$, $\lambda_{\text{max}}$  & . & 2.1\% (0.36), 8 & . & . & . & 3.9\% (0.25), 16 & . \\
& $\Delta\DALIStrong (\text{p-val})$, $\lambda_{\text{max}}$  & . & 2.7\% (0.28), 13 & . & . & . & 3.5\% (0.22), 9 & . \\
& $\Delta\MEXAtask (\text{p-val})$, $\lambda_{\text{max}}$  & . & 8.5\% (0.12), 23 & . & . & . & 1.6\% (0.30), 24 & . \\
\hline
\hline
\end{tabular}
}

\caption{$\Delta$ Alignment, p-val, $n_{\text{TS}}$ and $n_{\text{TF}}$, and $\lambda_\text{max}$ across the three NLU benchmarks in Llama3.1 8B. }
\label{tab:xlingalignment_stats}
\end{table*}

 \begin{figure*}[t!]

\begin{subfigure}{\textwidth}
    \centering
    \includegraphics[scale=0.5]{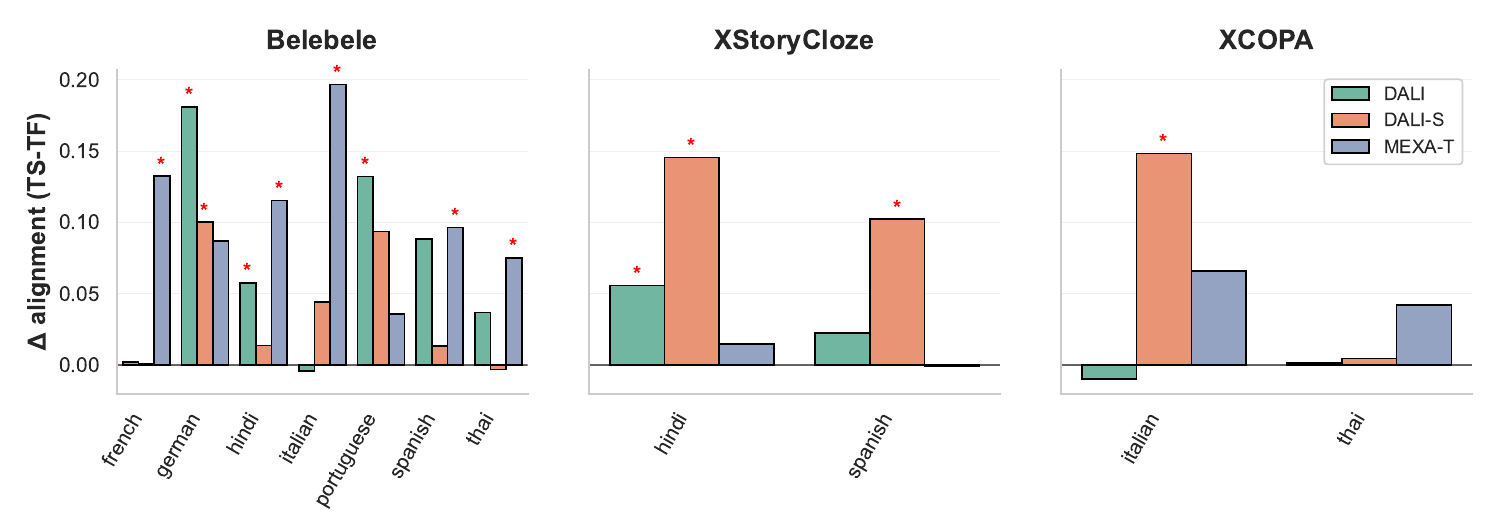}
    \caption{Llama3.1 8B it}
    \label{fig:alignmentresults_Llama3.18bit_appendix}
\end{subfigure}

\vspace{1em}

\begin{subfigure}{\textwidth}
    \centering
    \includegraphics[scale=0.5]{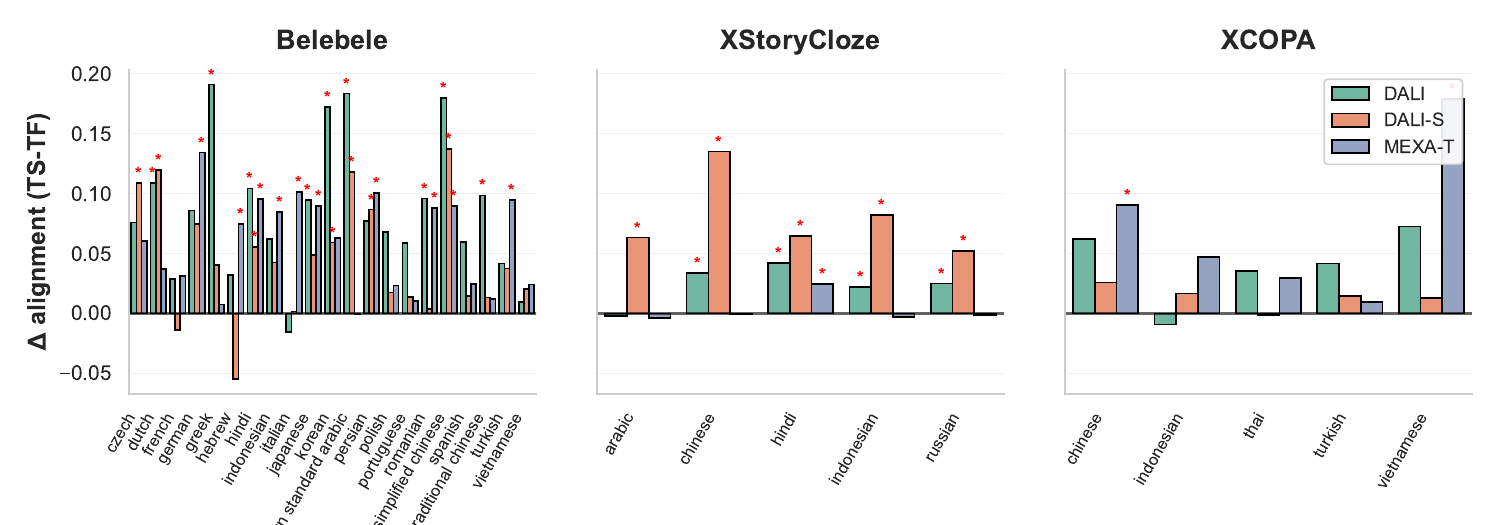}
    \caption{Aya23}
    \label{fig:alignmentresults_Aya23_appendix}
\end{subfigure}
\caption{$\Delta$ Alignment between TS and TF samples}
\label{fig:alignmentresults_appendix}
\end{figure*}

\begin{figure*}[!ht]
\centering
\setlength{\tabcolsep}{5pt} 
\begin{subfigure}[b]{\textwidth}
    \centering
    \includegraphics[width=0.8\textwidth]{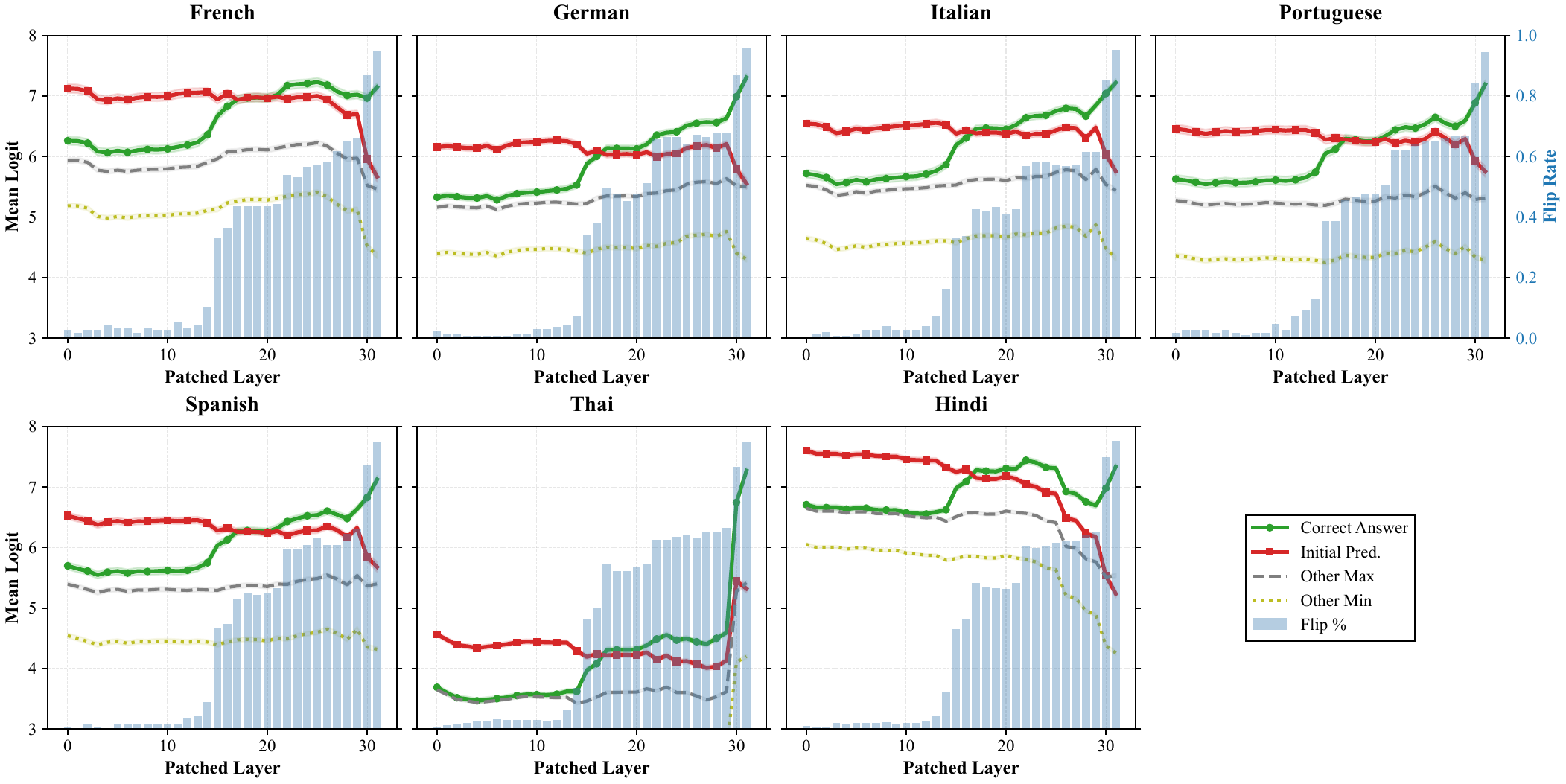}
    \caption{Last Token}
\label{fig:patching_results_belebele_lasttoken}
\end{subfigure}

\vspace{1em}

\begin{subfigure}[b]{\textwidth}
    \centering
    \includegraphics[width=0.8\textwidth]{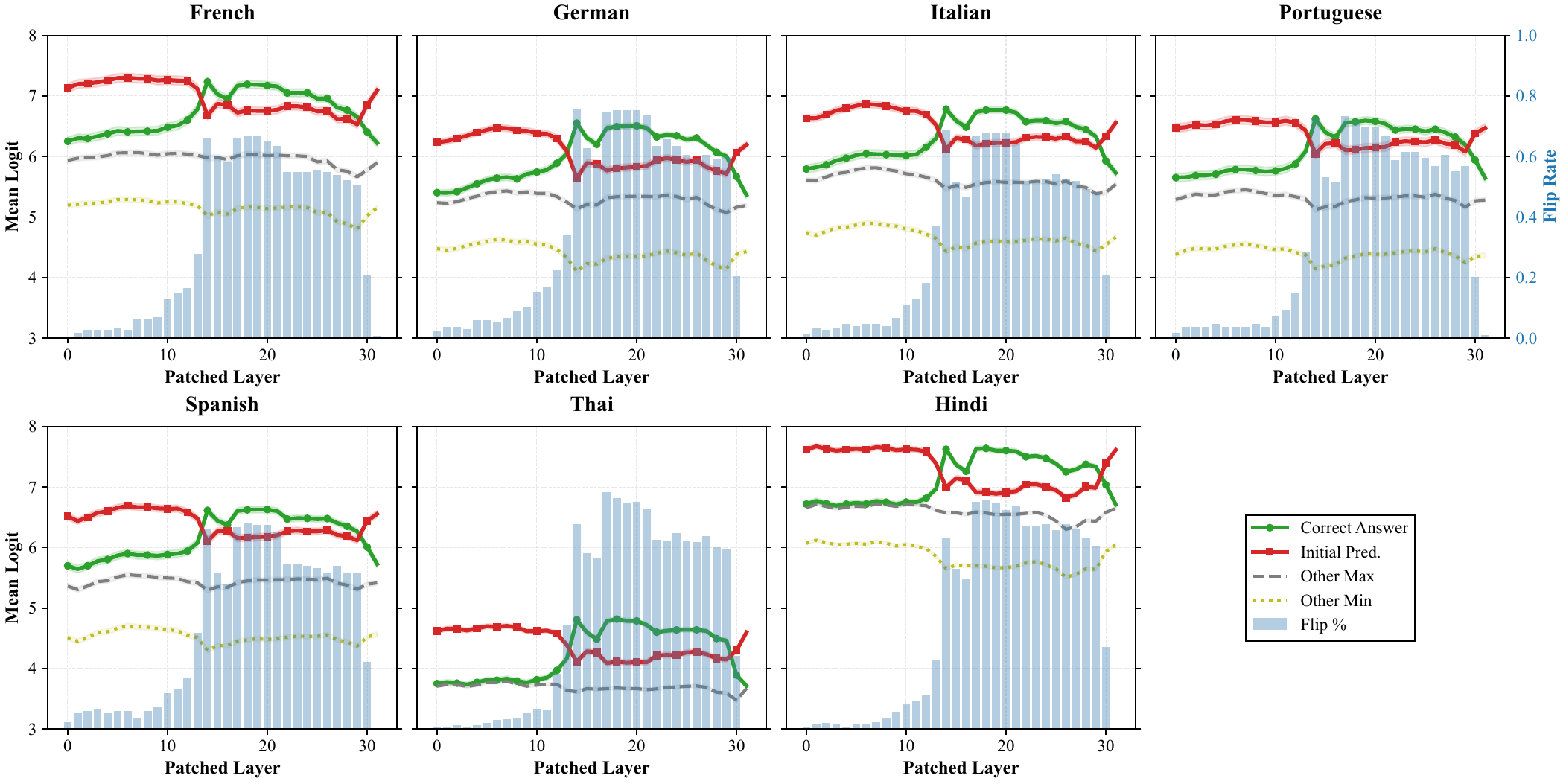}
    \caption{Penultimate Token}
\label{fig:patching_results_belebele_penultimatetoken}
\end{subfigure}

\caption{Patching results for the belebele benchmark across all languages in Llama 3.1 8B at last token (\subref{fig:patching_results_belebele_lasttoken}) and penultimate token (\subref{fig:patching_results_belebele_penultimatetoken}) respectively }
\label{fig:patching_results_belebele_appendix}
\end{figure*}

\begin{figure*}[!ht]
\centering
\setlength{\tabcolsep}{5pt} 
\begin{subfigure}[b]{\textwidth}
    \centering
    \includegraphics[width=0.7\textwidth]{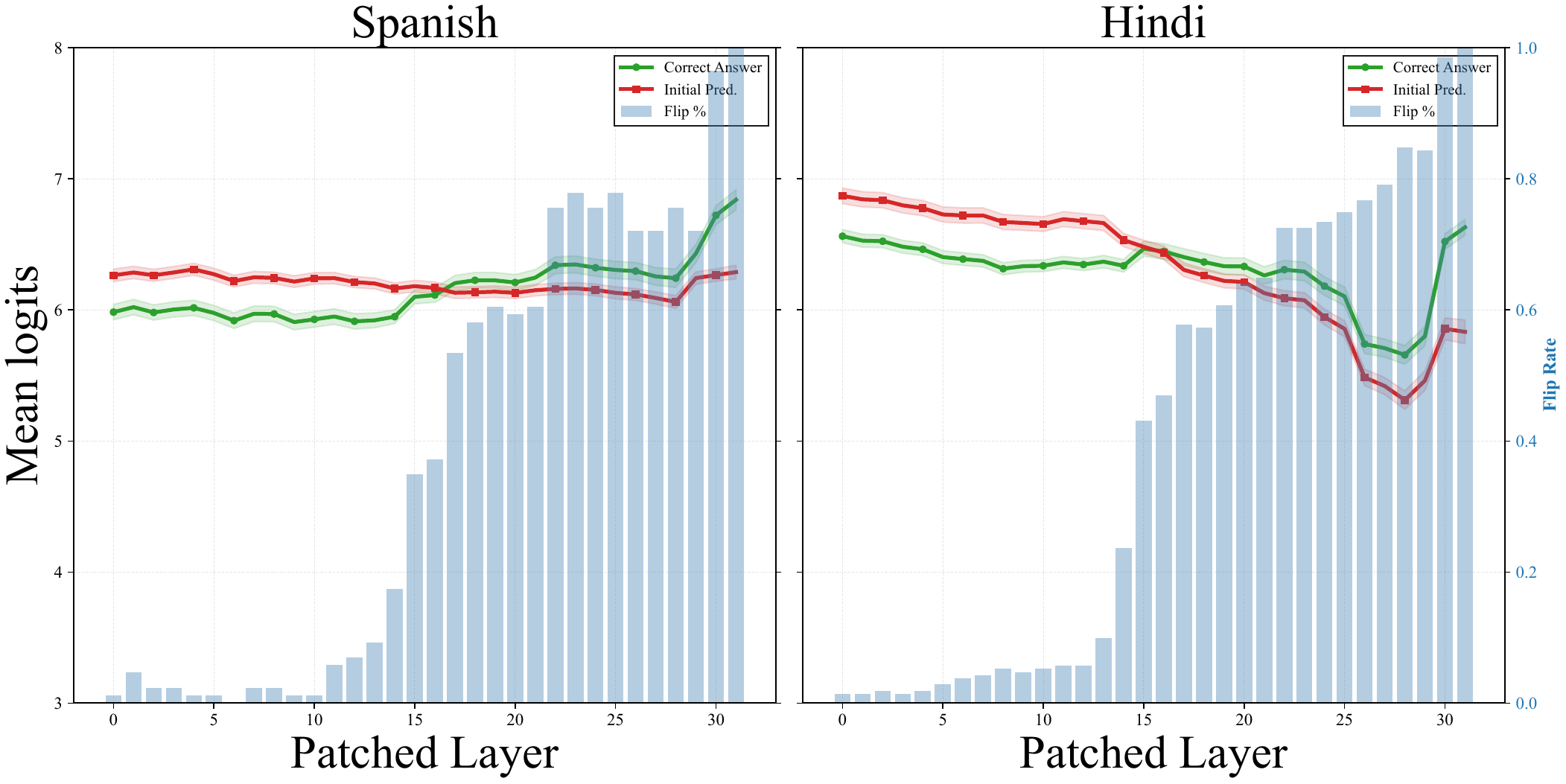}
    \caption{Last Token}
\label{fig:patching_results_xstorycloze_lasttoken}
\end{subfigure}

\vspace{1em}

\begin{subfigure}[b]{\textwidth}
    \centering
    \includegraphics[width=0.7\textwidth]{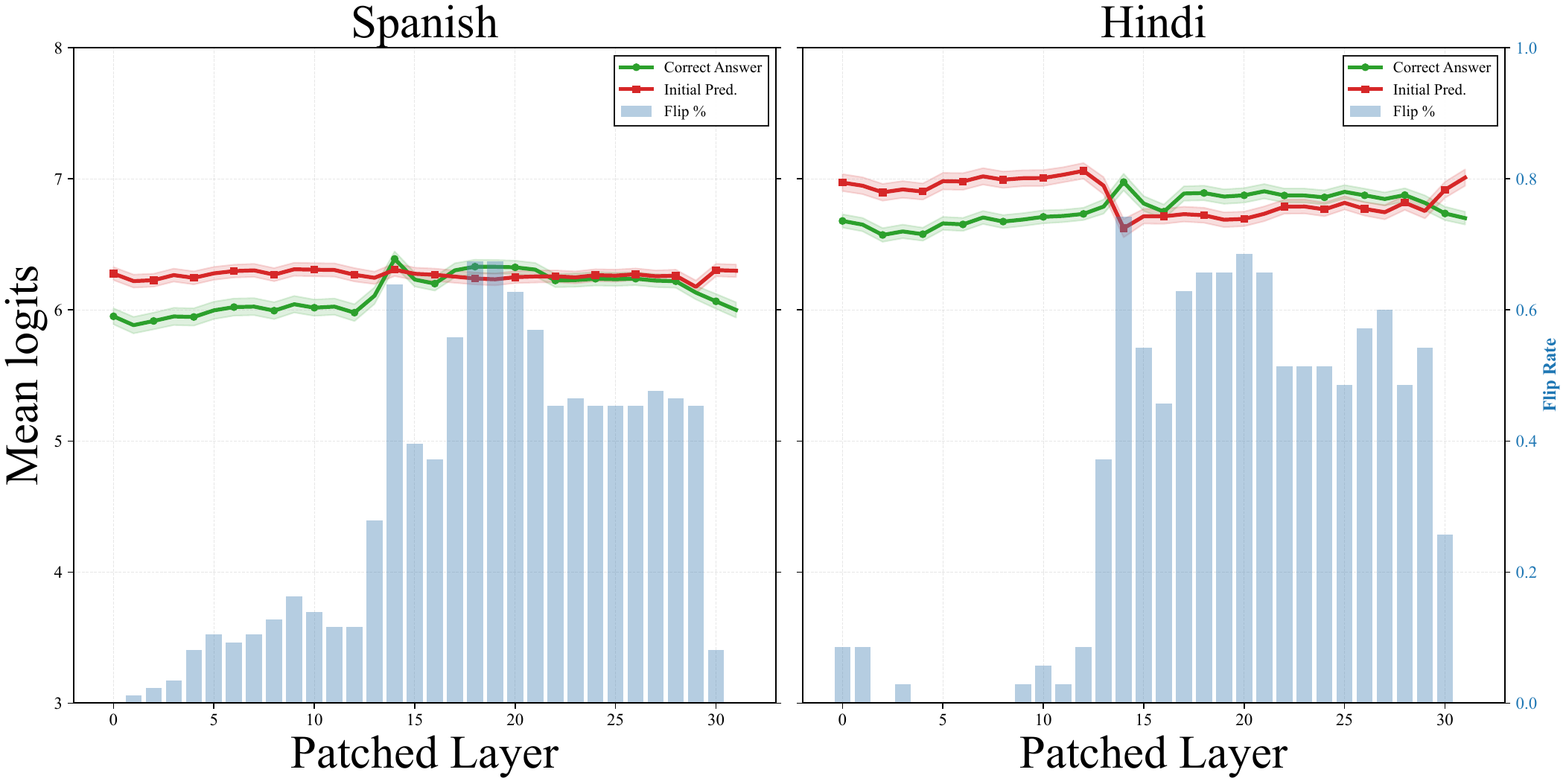}
    \caption{Penultimate Token}
\label{fig:patching_results_xstorycloze_penultimatetoken}
\end{subfigure}

\caption{Patching results for the xstorycloze benchmark across all languages in Llama 3.1 8B at last token (\subref{fig:patching_results_xstorycloze_lasttoken}) and penultimate token (\subref{fig:patching_results_xstorycloze_penultimatetoken}) respectively }
\label{fig:patching_results_xstorycloze_appendix}
\end{figure*}

\begin{figure*}[!ht]
\centering
\setlength{\tabcolsep}{5pt} 
\begin{subfigure}[b]{\textwidth}
    \centering
    \includegraphics[width=0.7\textwidth]{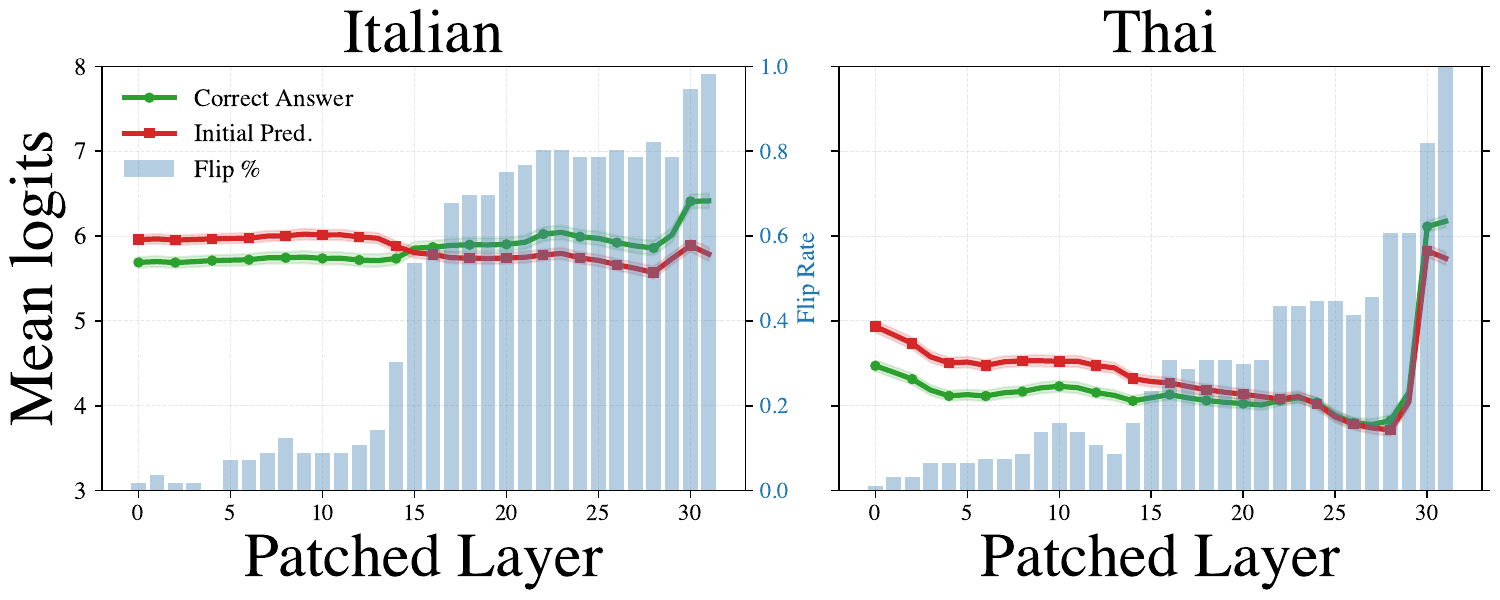}
    \caption{Last Token}
\label{fig:patching_results_xcopa_lasttoken}
\end{subfigure}

\vspace{1em}

\begin{subfigure}[b]{\textwidth}
    \centering
    \includegraphics[width=0.7\textwidth]{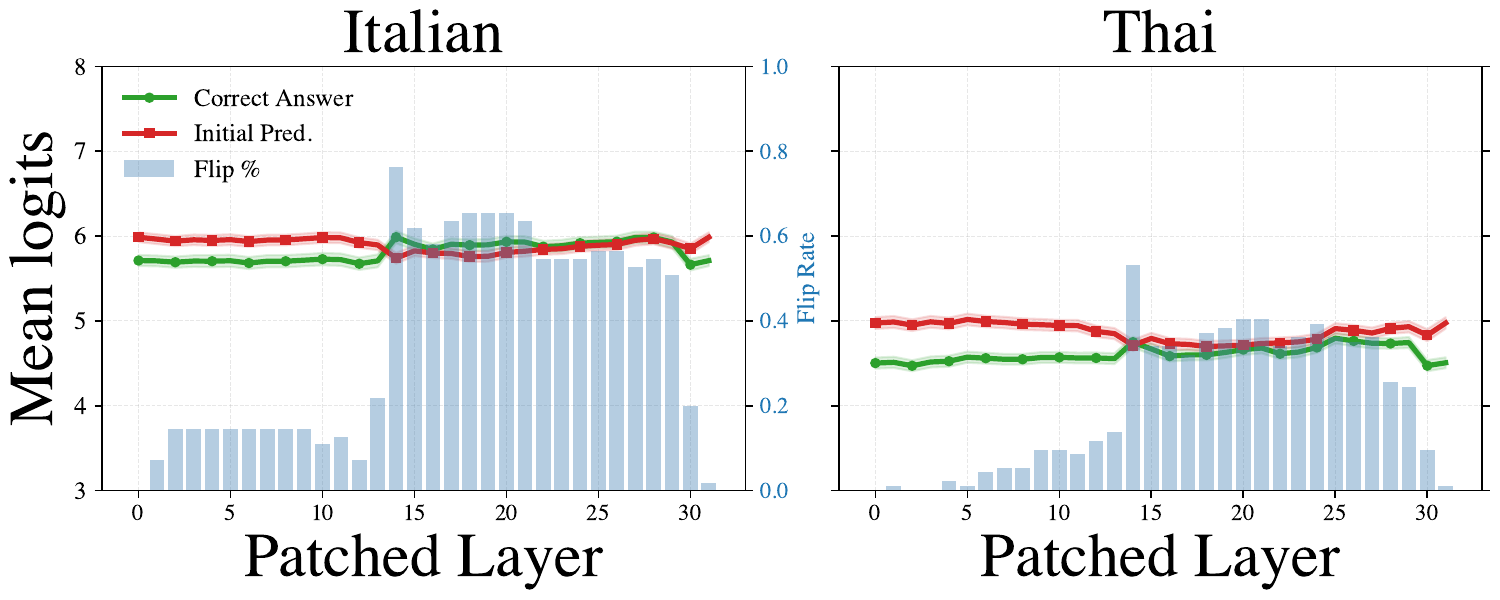}
    \caption{Penultimate Token}
\label{fig:patching_results_xcopa_penultimatetoken}
\end{subfigure}

\caption{Patching results for the xcopa benchmark across all languages in Llama 3.1 8B at last token (\subref{fig:patching_results_xcopa_lasttoken}) and penultimate token (\subref{fig:patching_results_xcopa_penultimatetoken}) respectively }
\label{fig:patching_results_xcopa_appendix}
\end{figure*}

 \begin{figure*}[t!]
    \centering
    \includegraphics[width=\linewidth]{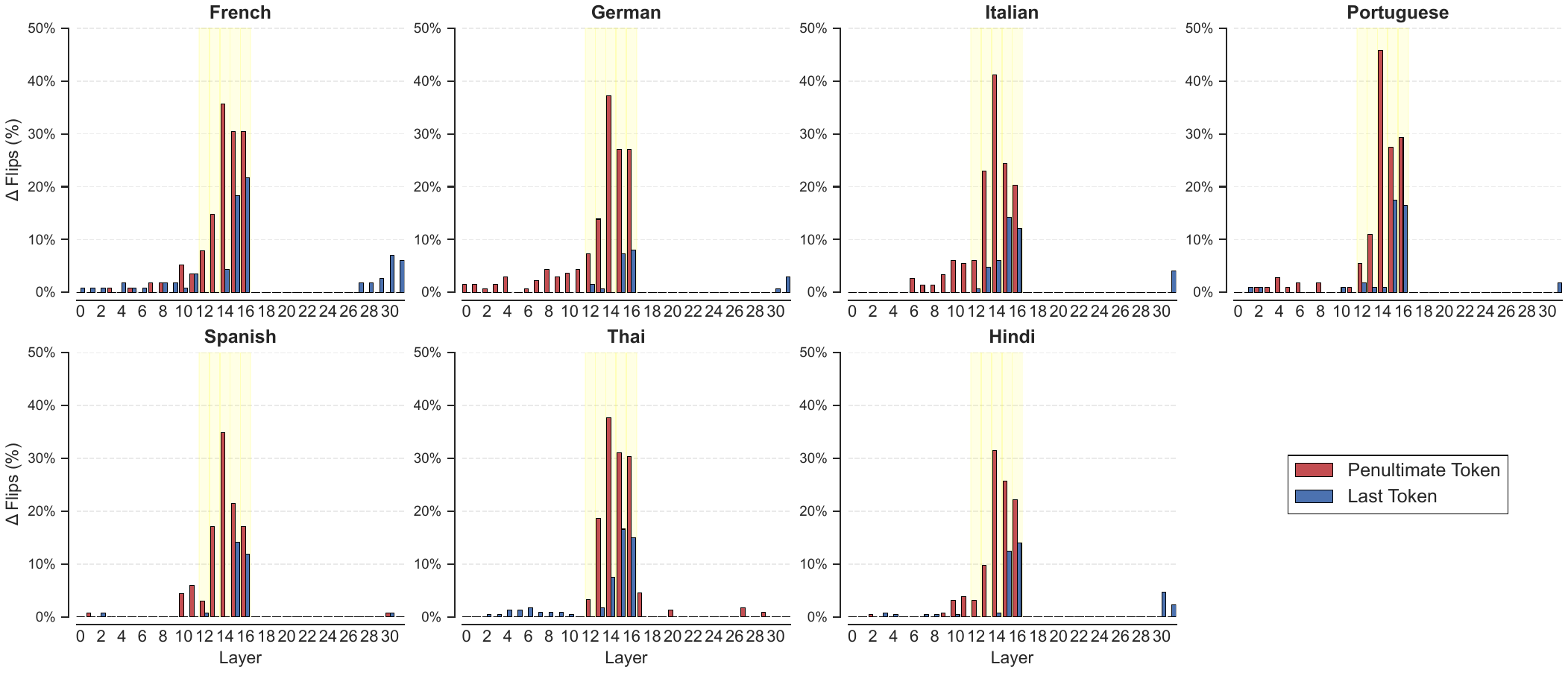}
    \caption{Control patching results for the belebele benchmark} \label{fig:control_exp_results}
\end{figure*}

 \begin{figure*}[t!]
    \centering
    \includegraphics[width=\linewidth]{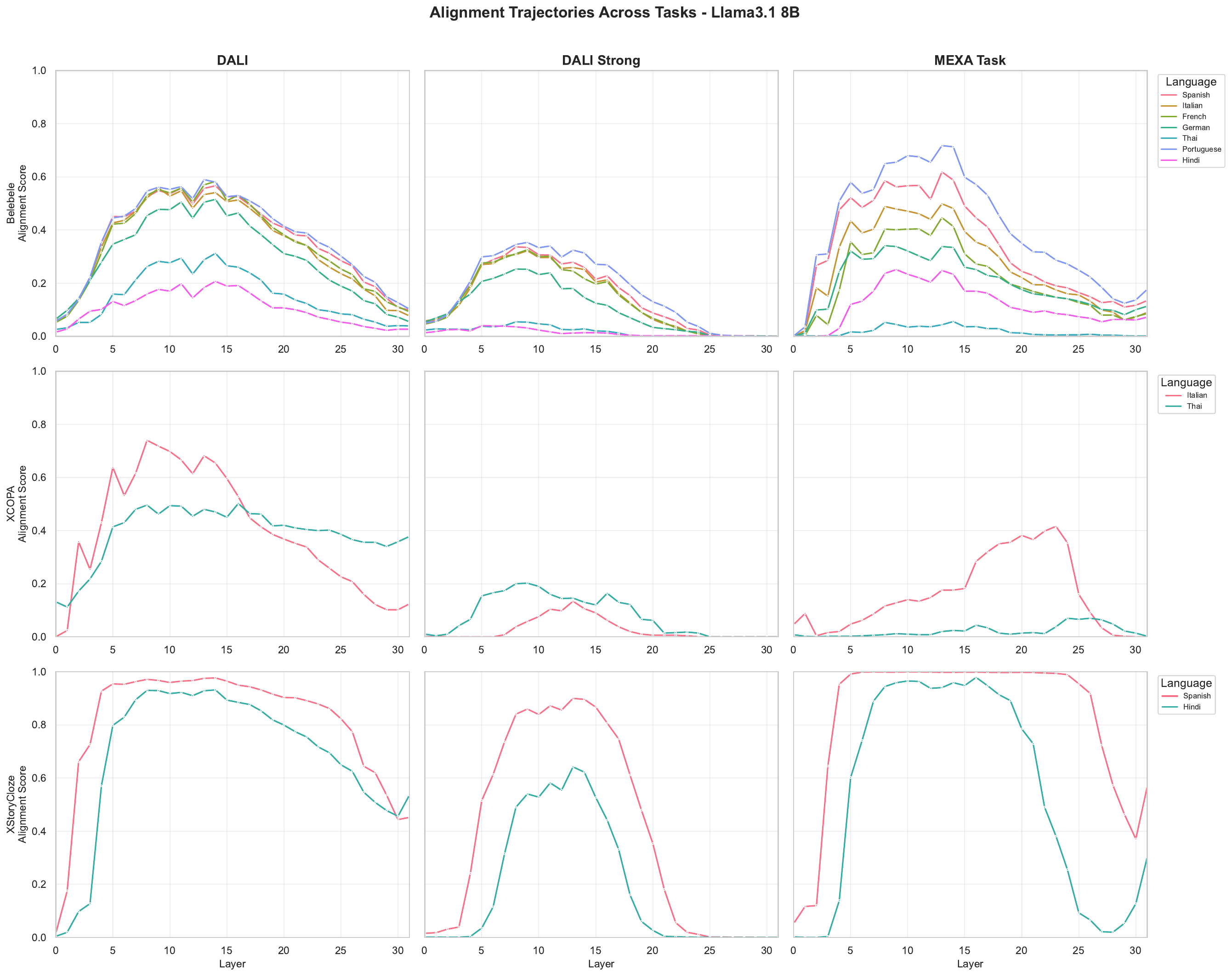}
    \caption{Alignment trajectories $\DALI$, $\DALIStrong$, $\MEXAtask$ by languages in Llama3.1-8B} \label{fig:alignment_trajectories}
\end{figure*}

\end{document}